\pdfoutput=1
\documentclass[11pt]{article}

\usepackage[preprint]{acl}

\usepackage{graphicx}
\usepackage{booktabs} % for professional tables

\usepackage{hyperref}

\usepackage{times}
\usepackage{latexsym}

\usepackage[T1]{fontenc}
\usepackage[utf8]{inputenc}

\usepackage{microtype}

\usepackage{inconsolata}

\usepackage{graphicx}
\usepackage{amsthm}
\usepackage{amsmath}
\usepackage[capitalize,noabbrev]{cleveref}
\theoremstyle{plain}
\newtheorem{theorem}{Theorem}[section]

\theoremstyle{definition}
\newtheorem{definition}[theorem]{Definition}

\theoremstyle{remark}

\usepackage[textsize=tiny]{todonotes}

\newcommand{\method}{\ensuremath{\textnormal{\textsc{SeaL}}}\xspace}
\newcommand{\cmethod}{\ensuremath{\textnormal{\textsc{SeaL-C}}}\xspace}

\usepackage{enumitem}
\usepackage{enumerate}
\usepackage{tabularx}
\usepackage{multirow}
\usepackage{url}
\usepackage{marvosym}
\usepackage{multirow}
\usepackage{algorithm}
\usepackage{algorithmic}
\renewcommand{\algorithmiccomment}[1]{// #1}
\usepackage{subcaption}
\renewcommand{\algorithmicrequire}{\textbf{Input:}}
\renewcommand{\algorithmicensure}{\textbf{Output:}}

\usepackage[utf8]{inputenc} % allow utf-8 input
\usepackage[T1]{fontenc}    % use 8-bit T1 fonts
\usepackage{url}            % simple URL typesetting
\usepackage{amsfonts}       % blackboard math symbols
\usepackage{nicefrac}       % compact symbols for 1/2, etc.
\usepackage{microtype}      % microtypography
\usepackage{colortbl}
\usepackage{tabularx}
\usepackage{makecell}
\usepackage{longtable}
\usepackage{float}
\usepackage{arydshln}
\usepackage{xcolor}
\usepackage{dutchcal}
\definecolor{mylightgreen}{rgb}{0.8, 1.0, 0.8}
\definecolor{mylightblue}{rgb}{0.68, 0.85, 0.9}

\definecolor{mycolor}{HTML}{00FF00}

\usepackage{amssymb}
\usepackage{mathtools}
\usepackage{siunitx}
\usepackage[capitalize,noabbrev]{cleveref}
\usepackage{enumitem}
\theoremstyle{plain}
\theoremstyle{definition}
\usepackage{longtable}
\usepackage{caption}
\usepackage[title,titletoc]{appendix}

\usepackage{bm} % for bold math
\usepackage{xspace} % for smart spacing
\usepackage{multirow} % for multirow command in tables
\usepackage{fontawesome}

\title{How Far are LLMs from Real Search? A Comprehensive Study on Efficiency, Completeness, and Inherent Capabilities}

\author{Minhua Lin$^{1}$\thanks{Work done during an internship at Amazon.}, Hui Liu$^{2}$, Xianfeng Tang$^2$, Jingying Zeng$^2$, \textbf{Zhenwei Dai}$^{2}$,\\ \textbf{Chen Luo}$^2$, \textbf{Zheng Li}$^2$, \textbf{Xiang Zhang}$^1$, \textbf{Qi He}$^{2}$, \textbf{Suhang Wang}$^{1}$   \\  
 $^{1}$The Pennsylvania State University $^{2}$Amazon\\
 \texttt{\{mfl5681,xzz89,szw494\}@psu.edu}\\ 
\texttt{\{liunhu,xianft,zejingyi,zwdai,cheluo,amzzhe\}@amazon.com} \\
}

\begin{document}
\maketitle

\begin{abstract}
% Search plays a fundamental role in problem-solving across various domains, with most real-world decision-making problems being solvable through systematic search. While traditional approaches suffer from efficiency issues, recent methods that rely solely on Large Language Models' (LLMs) intrinsic knowledge for search simulation prove unstable. Inspired by recent discussion in search and learning, we investigate the integration of LLMs into real search processes. First, we explore how LLMs can enhance search efficiency and propose a framework, \textbf{\underline{Se}arch vi\underline{a} \underline{L}earning (\method)}, for accurate and efficient search. Second, we study how to leverage LLMs to conduct complete and sound searches with 100\% accuracy efficiently. Finally, we explore whether LLMs can inherently develop search capabilities themselves. Our experimental results across three real-world planning tasks demonstrate that our framework can solve problems via complete but efficient search with 100\% accuracy using minimal model calls. Furthermore, we show that with proper guidance, LLMs can inherently conduct systematic searches, significantly improving their task-solving performance. These findings suggest promising directions for better empowering LLMs with search capabilities.

Search plays a fundamental role in problem-solving across various domains, with most real-world decision-making problems being solvable through systematic search. 
Drawing inspiration from recent discussions on search and learning, we systematically explore the complementary relationship between search and Large Language Models (LLMs) from three perspectives. 
First, we analyze how learning can enhance search efficiency and propose \textbf{\underline{Se}arch vi\underline{a} \underline{L}earning (\method)}, a framework that leverages LLMs for effective and efficient search. Second, we further extend \method to \cmethod to ensure rigorous completeness during search.
Our evaluation across three real-world planning tasks demonstrates that \method achieves near-perfect accuracy while reducing search spaces by up to 99.1\% compared to traditional approaches. 
Finally, we explore how far LLMs are from real search by investigating whether they can develop search capabilities independently. Our analysis reveals that while current LLMs struggle with efficient search in complex problems, incorporating systematic search strategies significantly enhances their problem-solving capabilities. These findings not only validate the effectiveness of our approach but also highlight the need for improving LLMs' search abilities for real-world applications.

% Search plays a fundamental role in problem-solving across various domains, with most real-world decision-making problems being solvable through systematic search. Drawing inspiration from recent discussions on search and learning, we systematically explore the complementary relationship between search and Large Language Models (LLMs) from three perspectives. First, we analyze how learning can enhance search efficiency and propose Search via Learning (SeaL), a framework that leverages LLMs for effective and efficient search. Second, we further extend SeaL to SeaL-C to ensure rigorous completeness during search. Our evaluation across three real-world planning tasks demonstrates that SeaL achieves near-perfect accuracy while reducing search spaces by up to 99.1\% compared to traditional approaches. Finally, we explore how far LLMs are from real search by investigating whether they can develop search capabilities independently. Our analysis reveals that while current LLMs struggle with efficient search in complex problems, incorporating systematic search strategies significantly enhances their problem-solving capabilities. These findings not only validate the effectiveness of our approach but also highlight the importance of improving LLMs' inherent search abilities for real-world applications.
\end{abstract}

\section{Introduction}
\label{sec:introduction}
Search lies at the heart of problem-solving, offering a systematic approach to explore solution spaces and find optimal answers. From everyday choices to complex strategic planning, virtually all real-world decision-making processes can be formulated and solved through systematic search strategies. This insight aligns with the recent discussions on search and learning~\cite{Sutton2019BitterLesson,snell2024scaling}, which emphasizes that the most effective problem-solving approaches combine systematic search with learning from experience.

Traditional search methods, such as brute-force searches, while theoretically complete, face challenges in systematically exploring large and complex search spaces.
They are designed to ensure that all possible states are considered, but the vastness of such spaces often necessitates exhaustive exploration, making systematic traversal impractical. Moreover, these methods lack the intuitive problem-solving abilities that humans naturally employ, such as recognizing promising solutions early or quickly abandoning unproductive paths.

Recent advances in Large Language Models (LLMs) and Large Reasoning Models (LRMs) have opened new possibilities for more human-like search approaches. Some recent LRMs such as OpenAI o1~\cite{openaio1card}, QwQ-32B~\cite{qwq-32b-preview}, and DeepSeek-R1~\cite{deepseekai2025deepseekr1} have demonstrated remarkable performance by incorporating LLM-guided search strategies. These approaches leverage LLMs' extensive knowledge bases and reasoning capabilities to guide the search process~\cite{wei2022chain,wang2023selfconsistency,snell2024scaling,yao2023tree,hao2023reasoning,wang2024litesearch}, attempting to mirror human-like intuition in problem-solving.

Despite the great success of these LLM-based simulated searches, a critical limitation remains: they rely heavily on models' intrinsic knowledge rather than combining it with systematic search strategies. When confronted with complex problems requiring multi-step reasoning or extensive exploration, these models often struggle to maintain consistent performance~\cite{wang2024q,wang2024multi,snell2024scaling}. Their reasoning can become unstable or incomplete, particularly when solutions require carefully exploring multiple solution paths or backtracking from dead ends—abilities that humans naturally employ in problem-solving.
% \suhang{cite???}.

Draw inspiration from recent discussions on search and learning~\cite{Sutton2019BitterLesson,snell2024scaling}, which emphasizes the great power of combining search and learning over human-centric approaches, in this paper, we systematically explore the integration and complementation of search and LLMs from three crucial perspectives: efficiency, completeness, and inherent search capabilities. 

Firstly, we explore how learning can benefit search. Specifically, we conduct a preliminary analysis to compare existing traditional and LLM-based search methods on a representative task, Game of 24, to investigate their problem-solving capabilities. Our experimental results reveal that the learning and reasoning capabilities of LLMs help solve simpler problems without extensive searching, reducing unnecessary state exploration and prioritizing promising states, which significantly shrink search spaces. 
Then, building upon these insights, we present \method, a framework that integrates learning into search algorithms to improve the search efficiency while maintaining completeness for enhancing the problem-solving capabilities of LLMs. 
We also introduce \cmethod, a variant that rigorously ensures search completeness while preserving efficiency through learning-guided complete state decomposition and two-phase ranking. 

We evaluate \method and \cmethod on three planning tasks: Game of 24, Mini Crosswords~\cite{yao2023tree} and Blocksworld~\cite{valmeekam2022large} using five representative LLMs, GPT-4o~\cite{gpt4ocard}, GPT-4o-mini~\cite{achiam2023gpt}, Qwen2.5-72B-Instruct~\cite{qwen2.5}, QwQ-32B-Preview~\cite{qwq-32b-preview} and DeepSeek-R1~\cite{deepseekai2025deepseekr1}. Our experimental results show that \method reaches almost perfect pass rates across almost all settings, while reducing search space 
by up to $99.1\%$ compared to traditional brute-force searches. And \cmethod is also proved to ensure rigorous completeness efficiently. These validate the effectiveness of \method in enabling complete and efficient search via learning. 

To this end, we {investigate how far LLMs are from real search} by investigating a reverse but natural problem: \textit{how search can benefit LLMs and whether LLMs can learn to search by themselves}. Specifically, we prompt LLMs to conduct searches solely relying on intrinsic knowledge or guided by \method's search strategies, respectively. Our analysis yields two significant insights. First, while search capabilities are crucial for LLMs' problem-solving effectiveness, current LLMs/LRMs exhibit inefficient search behaviors, requiring extensive sampling to achieve satisfactory performance. Second, incorporating \method's search strategies into LLM prompts demonstrably enhances their problem-solving capabilities. These findings not only underscore the critical role of search in enhancing LLMs' reasoning and learning capabilities but also validate the effectiveness of \method's search strategies, motivating us to improve LLMs' self-search capabilities in the future works. Our \textbf{main contributions} are: 
\begin{itemize}[leftmargin=*,noitemsep, topsep=0pt]
    \item (\textbf{Analysis}) Inspired by the principles of search and learning, we conduct a systematic exploration of how learning benefits search, demonstrating that LLMs can reduce unnecessary state exploration and improve efficiency by prioritizing promising states. Additionally, we explore how search benefits LLMs, revealing the importance of teaching LLMs to efficiently search for solving complex problems.
    \item (\textbf{Methodology}) We propose \method, a framework that integrates learning into search for efficiency and completeness, and \cmethod, a variant to ensure rigorous completeness.
    \item (\textbf{Experiments}) We evaluate \method and \cmethod across diverse real-world tasks, demonstrating their effectiveness in achieving efficient and complete search. We also reveal that existing LLMs cannot perform efficient and effective search regarding complex tasks, which is a desired fundamental ability for LLMs to be applied to real-world decision making tasks.
\end{itemize}

\section{Related Works}
\label{sec:related_works}
\textbf{LLM-based Search Methods}.  
Recent advancements in test-time compute scaling have sparked growing interest in methods that enable LLMs to simulate search processes and ``think longer'' instead of directly generating answers in one pass. Several works~\cite{wang2023selfconsistency,hao2023reasoning,feng2023alphazerolike,yao2023tree,zhao2024large,besta2024got,wang2024litesearch,snell2024scaling} adopt such approaches to enhance LLMs' problem-solving capabilities. 
Despite their innovation, these methods rely solely on LLMs' intrinsic knowledge, often leading to unstable performance due to limitations in LLMs' reasoning capabilities. In contrast, our proposed \method integrates LLMs with traditional search strategies, ensuring both completeness and efficiency in solving decision-making tasks. More LLM-based search methods are reviewed in Appendix~\ref{appendix:related_works_LLM_search}.

\noindent \textbf{Traditional Search Methods}.  
Inspired by the recent discussion about search and learning~\cite{Sutton2019BitterLesson}, which underscores the enduring value of general-purpose strategies that scale with computational power, we investigate how to better integrate search with learning to leverage the strengths of both paradigms. Our \method and \cmethod draw inspiration from traditional search while incorporating LLM-guided reasoning to reduce search space, thereby significantly enhancing efficiency without sacrificing completeness. More details of the related works are in Appendix~\ref{appendix:related_works}. 
\section{How Learning can Benefit Search}

Search and learning represent two fundamental approaches to problem-solving. Traditional search methods offer systematic exploration with guaranteed completeness, while learning-based approaches leverage pattern recognition to identify promising solutions quickly. 
In this section, we conduct a preliminary analysis to systematically explore {existing search algorithms and} the synergies between learning and search. Building on these insights, we introduce \method, a framework that integrates learning into search algorithms to reduce unnecessary exploration, prioritize promising paths, and ultimately maintain reliable and efficient performance as problems grow in complexity.

\subsection{Experimental Setup}
\label{sec:preliminary_setup}
\textbf{Task Setup}. To investigate the impact of learning on search, we use \textit{the Game of 24} as a representative task. This task can be solved using traditional search algorithms and is also widely adopted for evaluating LLMs' planning abilities. Details about the task are provided in Appendix~\ref{appendix:tasks}. Following the setting in \citet{yao2023tree}, we select \num{100} problems indexed as $900-999$ for our experiments.

\noindent \textbf{Baselines.} For our preliminary analysis, we evaluate three LLM-based simulated search algorithms: \textsc{Majority Vote} \cite{wang2023selfconsistency}, \textsc{Best-of-N} \cite{snell2024scaling}, and \textsc{Beam Search} \cite{yao2023tree}. Additionally, we include \textsc{Vanilla CoT} (Chain-of-Thought)~\cite{wei2022chain} as a reference. To evaluate search efficiency, we also consider two traditional brute-force search methods, Depth-First Search (\textsc{DFS}) and Breadth-First Search (\textsc{BFS}), along with their pruning variants, \textsc{DFS-Prune} and \textsc{BFS-Prune}. These variants improve efficiency by avoiding exploration of previously visited states. Finally, an \textsc{Exhaustive Search} is included as a baseline for comprehensive comparison.
Detailed implementation details for all baselines are provided in Appendix~\ref{appendix:baselines}.

\noindent \textbf{Evaluation Metrics}  We evaluate performance using two primary metrics:
(i) pass rates (PR) across games per difficulty level, measuring solution quality, and
(ii) search steps (SS), measuring exploration efficiency by counting traversed states in the search space $\mathcal{S}$.
Complete metric definitions are available in Appendix~\ref{appendix:evaluation_metrics}.

\begin{figure}[t]
    \small
    \centering
    \begin{subfigure}{0.49\linewidth}
        \includegraphics[width=0.98\linewidth]{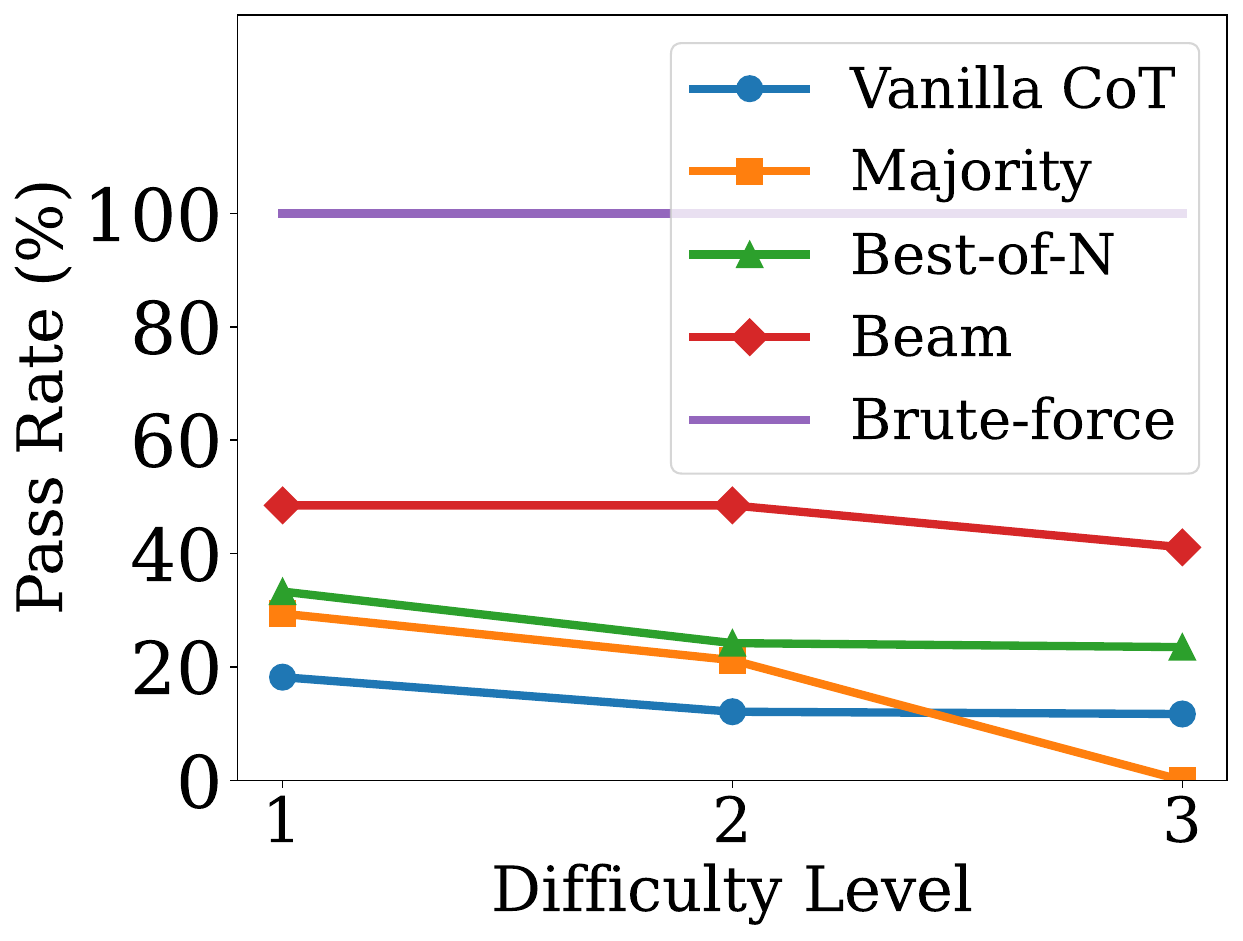}
        \vskip -0.5em
        \caption{Pass Rate}
    \end{subfigure}
    \begin{subfigure}{0.49\linewidth}
        \includegraphics[width=0.98\linewidth]{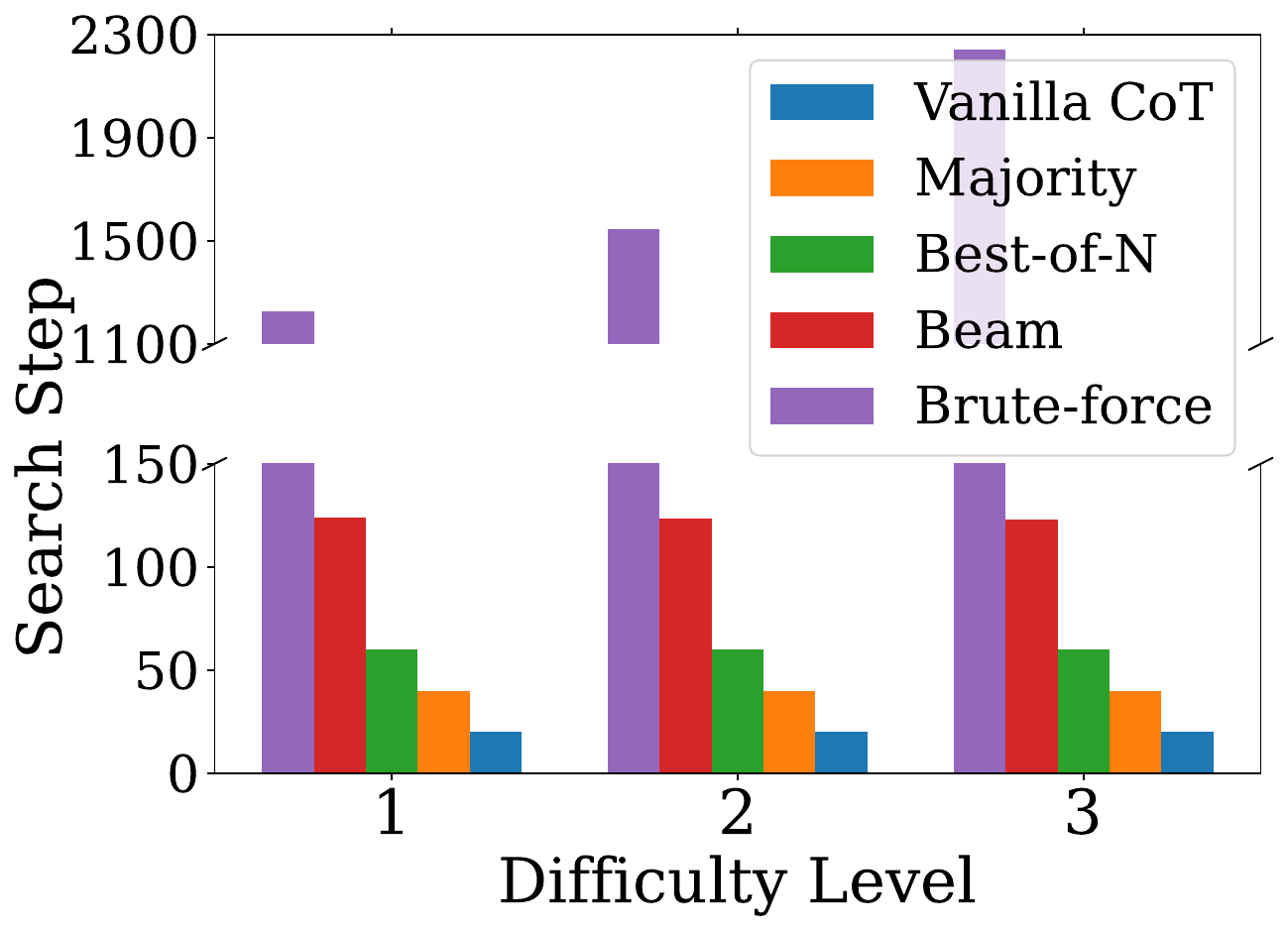}
        \vskip -0.5em
        \caption{Search Step}
    \end{subfigure}
    \vskip -1em
    \caption{PR (\%) and SS of existing searches across various problem difficulties using GPT-4o-mini. 
    % \hui{for Figure 1(b), can we plot a figure with y axis break. Otherwise it's hard to read.}
    }
    \vskip -1em
    \label{fig:preliminary_difficulty_level_group}
\end{figure}

\subsection{Analysis: How Learning can Benefit Search}
\label{sec:preliminary_results_analysis}
To better understand how learning can enhance search, we divide the 100 problems from the Game of 24 into three difficulty levels based on human success rates\footnote{{https://www.4nums.com/game/difficulties/}}. Other experimental settings follow those in Sec.~\ref{sec:preliminary_setup}.
Our experimental results, presented in Fig.~\ref{fig:preliminary_difficulty_level_group}, reveal several key findings regarding the performance of existing search methods. Full analyses are in Appendix~\ref{appendix:preliminary_results_analysis_appendix}.

\noindent \textbf{Obs. 1: LLMs Perform Better for Simpler Problems}. 
As in Fig.~\ref{fig:preliminary_difficulty_level_group}(a), \textsc{Vanilla CoT} achieve pass rates of $\num{18.2}\%$, $\num{12.1}\%$, and $\num{11.7}\%$ for problems with difficulty levels $1$, $2$, and $3$, respectively. This pattern suggests that LLMs excel at direct problem-solving for simpler cases, indicating their potential for single-step solutions rather than requiring iterative approaches. More examples are in Fig.~\ref{fig:obs1_example}.

\noindent \textbf{Obs. 2: Learning-Based Pruning Has Precision-Coverage Trade-offs}. According to Fig.~\ref{fig:preliminary_difficulty_level_group}(a) and~\ref{fig:preliminary_difficulty_level_group}(b), \textsc{Beam Search} achieves the superior performance in pass rate than other LLM-based search methods, and significantly saves search steps than \textsc{DFS}. To deeply analyze this observation, an example of \textsc{Beam Search} in Fig.~\ref{fig:obs2_example} shows that it only relies on LLMs to generate a few of possible next steps instead of decomposing all possible states in brute-force searches in each time of generating next intermediate steps. However, it is possible that some valid states are overlooked during the intermediate step generation, leading to failure in solving tasks and reflecting a trade-off between efficiency and completeness in search.

\noindent \textbf{Obs. 3: Learning Provides Adaptive Search Guidance}. 
Fig.~\ref{fig:obs3_example} is an example of \textsc{Beam Search} employing an LLM verifier to assess intermediate states' progress toward the goal, prioritizing exploration of valid states and avoiding further exploration on other states. Moreover,
as in Fig.~\ref{fig:preliminary_difficulty_level_group}(a), the LLM-based methods using more search steps generally perform better. This is because these methods allocate more search budget in verifying results via LLMs, further demonstrating LLMs' capability as dynamic evaluators to leverage learned knowledge to identify promising states and guide the search process.

These findings suggest that learning can substantially improve search efficiency when properly integrated into the search process. However, they also highlight the need for careful mechanism design to balance the benefits of learning-based pruning with the completeness guarantees of traditional search.

\begin{table}[t]
\caption{PR (\%) and SS on the Game of 24 using Qwen2.5-72B-Instruct. For LLM-based methods, we additionally present the reduction ratio of SS compared to \textsc{Brute-force (DFS)}.
}
\vspace*{-0.5em}
\centering
\small
\tabcolsep 3.5pt
\renewcommand\arraystretch{1.0}
\begin{tabular}{llccc}
\toprule
\multirow{2}{*}{} &
\multirow{1}{*}{Method} &
\multicolumn{1}{c}{PR (\%)} &
\multicolumn{1}{c}{SS} \\
\midrule
&\textsc{Exhaustive Search}&\num{100} &\num{12928}\\
\addlinespace[0.1em]\hdashline\addlinespace[0.1em]
&\textsc{Brute-force (DFS)}&\num{100} &\num{1623} \\
&\textsc{Brute-force (BFS)}&\num{100} &\num{3429} \\
&\textsc{Brute-force (DFS-Prune)}&\num{100} &\num{1385}\\
&\textsc{Brute-force (BFS-Prune)}&\num{100} &\num{1306}\\
\midrule
&\textsc{Vanilla CoT} &\num{17}   &\num{1} ({$\downarrow$} 99.9\%) \\
&\textsc{Majority vote} &\num{23}  &\num{10} ({$\downarrow$} 99.3\%) \\
&\textsc{Best-of-N}&\num{27}  &\num{20} ({$\downarrow$} 98.7\%)\\
% \rowcolor{gray!30} &\lmmweb\textsubscript{Oracle}
&\textsc{Beam search}
&\num{35} &\num{124} ({$\downarrow$} 92.4\%)\\
\bottomrule
\end{tabular}
\vspace{-1em}
\label{tab:preliminary_24game_qwen}
\end{table}
\subsection{Analysis: Existing Search Algorithms}
\label{sec:analysis_existing_search}
While existing search algorithms show significant promise, they still encounter notable challenges. 
To better illustrate their limitations, we conduct an analysis following the setup in Sec.\ref{sec:preliminary_setup} by applying existing search algorithms to the Game of 24 task.
The results are reported in Table~\ref{tab:preliminary_24game_qwen}, revealing the following insights: \textbf{(i)} Traditional brute-force algorithms achieve perfect $100\%$ accuracy but require up to \num{3429} search steps, demonstrating their inefficiency in searching for answers.
\textbf{(ii)} While LLM-based methods significantly reduce the search space (at least $92.4\%$ fewer steps than \textsc{DFS}), their pass rates peak at only $35\%$, highlighting their instability compared to traditional searches. Full analyses are in Appendix~\ref{appendix:analysis_existing_search_appendix}

These findings underscore the necessity of a framework that combines the completeness of systematic search with the efficiency of learning-based approaches. This motivates the development of a more effective search-and-learning methodology to address these limitations.

\begin{figure}[t]
    \small
    \centering
    
        \includegraphics[width=0.99\linewidth]{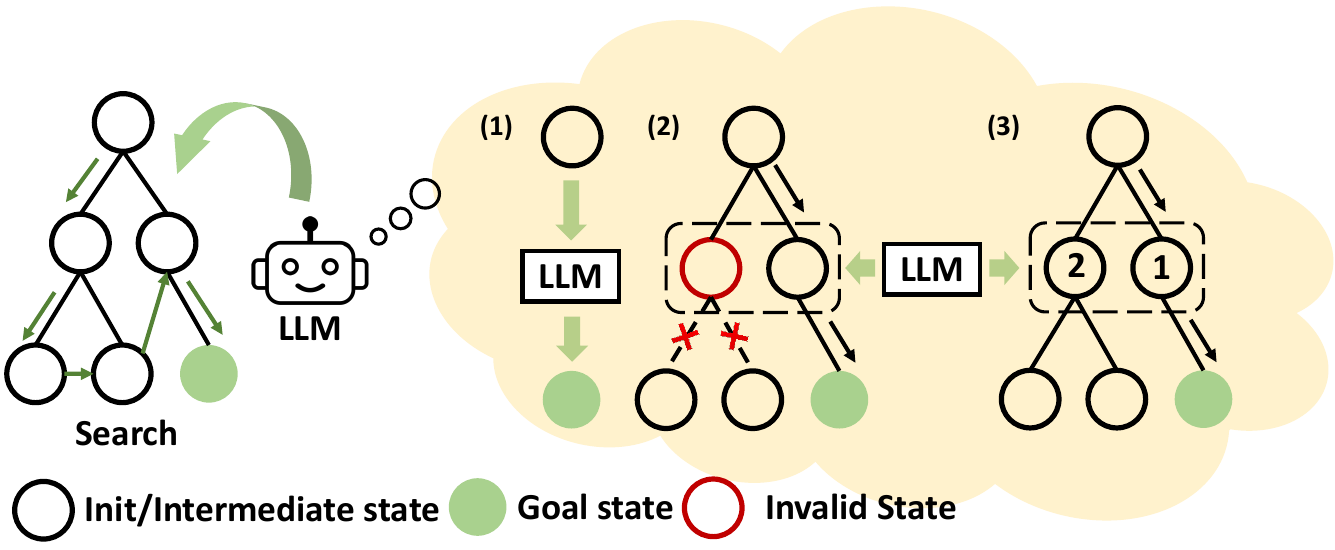}
        \vskip -1em
    % \vskip -1.em
    \caption{\method intergrating learning into search with LLMs: \textbf{(1)} Direct solution generation, \textbf{(2)} State validity checking, \textbf{(3)} Learning-guided state ranking.}
    \vskip -1.0em
    \label{fig:Framework_figure}
\end{figure}
\subsection{\method: Search via Learning with LLMs}
\label{sec:seal_method}
\noindent \textbf{Notations}.
We focus on solving planning problem following the essence of solving real-world decision-making problems via search: starting from an initial state, envisioning possible actions, and systematically working toward a goal state. 
Specifically, a planning problem is formally defined as a tuple $P=\langle\mathcal{S}, s^{init},s^{goal},\mathcal{A},f\rangle$. Here, $\mathcal{S}$ represents a finite and discrete set of states describing the world (i.e., state space). $s^{\text{init}}, s^{\text{goal}} \in \mathcal{S}$ denote the initial and goal world states, respectively. $\mathcal{A}=\{a_1, a_2, \ldots,\}$ represents the set of possible actions, and $f(s,a_i)=s'$ is a transition function mapping a state and action to a resulting state. A solution to problem $P$ is a sequence of actions $\langle a_1,a_2,\ldots\rangle$ that transforms $s^{\text{init}}$ into $s^{\text{goal}}$.

Building upon the above insights, we introduce \method, a framework that systematically integrates learning capabilities into search processes to emulate human problem-solving strategies. The goal of \method is on enhancing this natural problem-solving process by combining the systematic nature of search with the learning capabilities of LLMs.
\method consists of four components to enhance different aspects of search processes through learning:

\noindent \textbf{Direct Solution Generation}. Motivated by the insight from Obs. 1 that learning excels at solving simpler problems directly, we begin each step by attempting a direct solution, akin to how humans first try to solve problems in one step:
\begin{equation}
\small
r^{cur} = M(p_{\text{solve}}(s^{cur}))
\end{equation}
where $M$ represents the backbone LLM and $p_{\text{solve}}$ denotes the solution generation prompt. A verifier $f$ validates the generated solution. More details of $p_{solve}$ are in Appendix~\ref{appendix:prompt_templates_direct_solution_generation}.  This mechanism leverages LLMs' strength in handling simpler cases, potentially bypassing the need for extensive search when direct solutions are viable. Note that, however, the verifier $f$ may not always be available in certain scenarios. To ensure the applicability of \method in such cases, we provide further discussions in Appendix~\ref{appendix:impact_verifier_search}.

\noindent \textbf{State Decomposition}. For more complex problems that cannot be solved directly, we decompose the current state $s^{cur}$ into subproblems, mirroring how humans break down challenging tasks:
\begin{equation}
\label{eq:state_decomposition}
\small
    S^{next} = D(s^{cur}),
\end{equation}
where $D$ is the decomposition function generating substates $S^{next} = \{s^{next}_1,s^{next}_2,\ldots\}$. While LLMs can suggest promising decompositions~\cite{yao2023tree}, Obs. 2 reveals the risk of overlooking valid paths. Thus, we maintain the ability to systematically enumerate substates when needed, balancing between efficient exploration of valid paths and thorough coverage of the state space $\mathcal{S}$.

\noindent \textbf{State Validity Checking}. 
During search, it is inevitable that we may explore some invalid states, especially if we decompose into all possible substates. Exploring and expanding these invalid substates too much will significantly increase the search space. Thus, inspired from Obs. 3, we leverage LLMs to assess the potential of substates toward goals, similar to how humans quickly judge if it is worthy to continue exploration:
\begin{equation}
\small
% \begin{aligned}
    c^{next} = M(p_c(S^{next})),
% \end{aligned}
\end{equation}
where $p_c$ prompts the LLM $M$ to evaluate state validity. Additional examples $p_{c}$ are in Appendix~\ref{appendix:prompt_templates_state_validity_checking}. In this paper, we opt for binary decisions ($c^i_{k}$) about whether to continue exploration on the substates $S^{next}$, ensuring efficient pruning on invaid states while maintaining search on valid states.

\noindent \textbf{Learning-guided Ranking}
After validating substates, we prioritize paths similar to how humans naturally focus their attention on the most promising approaches. Leveraging Obs. 3, we employ LLMs to compute a priority score of each state:
\begin{equation}
\small
% \begin{aligned}
v(s^{cur}) = M(p_v(s^{cur}))
% \end{aligned}
\end{equation}
where $v(s^{cur})$ denotes the LLM-estimated value using prompt $p_v$, indicating how likely the current state leads to the goal state $s^{goal}$. States with higher scores are prioritized in the exploration queue, enabling efficient traversal of promising solution paths. Details of $p_{v}$ are in Appendix~\ref{appendix:prompt_templates_learning_guided_ranking}.

\noindent \textbf{Overall Alogirhtm}
Combining these components together into search, we implement the framework of \method, which is shown in Fig.~\ref{fig:Framework_figure}. The full algorithm is in Appendix~\ref{appendix:algorithm_full}. Our search strategy focuses on exploring promising paths deeply before considering alternatives, similar to how humans naturally approach problem-solving. This design allows \method to scale to complex problems effectively and efficiently through its combination of learning-guided intuition and systematic exploration.

\section{Toward Complete Search via Learning}
{While \method effectively balances efficiency and effectiveness, our observation indicates potential compromises in solution accuracy. In high-stakes domains such as autonomous driving~\cite{mao2023gpt} and pandemic response planning~\cite{du2024advancing}, ensuring completeness is fundamental to search algorithms, as overlooking any viable solution can have severe consequences. This motivates us to develop a search framework that rigorously ensures completeness. Next, we formalize search completeness and introduce \cmethod, an enhanced variant of \method{} that integrates efficient learning-guided exploration with formal guarantees of completeness.}

% \vspace{-1em}
\subsection{Formalizing Search Completeness}
\label{subsec:theoretical_foundations}
We begin by formalizing search completeness--the guarantee of finding a solution when one exists:
\begin{definition}[Search Completeness]
A search algorithm is complete if and only if for any initial state $s^{init} \in \mathcal{S}$ and goal state $s^{goal}$, whenever there exists a valid solution path $P = (s^0, ..., s^n)$ where $s^n \in s^{goal}$, the algorithm is guaranteed to find it.
\end{definition}

\subsection{\cmethod: Achieving Search Completeness}
\label{subsec:method_cs}
Building on this definition, we first analyze potential completeness compromises in \method, then present \cmethod's mechanisms for ensuring rigorous completeness. The full algorithm of \cmethod is shown in Alg.~\ref{alg:Framwork_completeness}.

\noindent \textbf{How Can Completeness Be Compromised}?
According to Sec.\ref{sec:seal_method}, \method uses LLMs for state decomposition and validity checking, inspired by Obs. 2 and 3. However, our analysis in Sec.\ref{sec:preliminary_results_analysis} reveals that they may inadvertently ignore valid states during decomposition or prematurely terminate exploration of valid paths, compromising completeness.

\noindent \textbf{Learning-Guided Complete State Decomposition}. To ensure completeness while maintaining efficiency, \cmethod employs a learning-guided complete state decomposition strategy by combining learning-based prioritization with a fallback mechanism for exhaustive state expansion:
\begin{equation}
% \begin{aligned}
\small
{S}^{next} = M\big(p_d(s^{cur})\big) || \big(D(s^{cur})\setminus M(p_d(s^{cur}))\big)
% \end{aligned}
\end{equation}
where $||$ denotes ordered concatenation, ensuring LLM-generated states $M(p_{{d}}(s))$ are explored first. $D(s)$ is the complete state decomposition function from Eq.~(\ref{eq:state_decomposition}). 
% $\setminus$ is the set difference, retaining the rest states in $D(s)$ not proposed by LLMs. 
This approach prioritizes exploration of likely valid states while ensuring no potential solution is overlooked, guaranteeing completeness while benefiting from learning-guided efficiency.

\noindent \textbf{Two-phase Ranking}. 
To further improve efficiency, \cmethod{} introduces a two-phase ranking strategy for $S^{next}$. Instead of ranking all states at once, it first ranks and explores the LLM-generated states $M(p_d(s^{cur}))$, which are more likely to reach $s^{goal}$. Only when no solution is found does the algorithm proceed to rank and explore the remaining states from $D(s^{cur})$, which significantly reduces the search space by avoiding unnecessary ranking of supplementary states.

\section{Can LLMs Learn to Search by Themselves?}
\label{sec:llm_self_search}
Having integrated learning to enhance search, a reverse but natural question emerges: 
% \textit{Can search strategies improve LLMs' reasoning capabilities?}
\textit{Can LLMs execute search autonomously to improve the reasoning capabilities?}
Recent advances in scaling test-time computation suggest that systematic exploration could enhance LLMs' problem-solving abilities. While models like QwQ-32B~\cite{qwq-32b-preview} show promising reasoning capabilities, they struggle with focused problem-solving, often producing unfocused, recursive outputs. This motivates us to explore how search benefits LLMs and whether LLMs possess the potential for self search.

To systematically investigate LLMs' self-search capabilities, we consider two types of prompts: high-level self-search, which relies solely on LLMs' internal knowledge, and low-level self-search, which explicitly encodes \method's search strategies into the prompts. Further details and illustrative examples are in Appendix~\ref{appendix:more_details_self_search}, with additional discussions and analyses provided in Sec.~\ref{sec:experiment_self_search}.

\section{Experiments}
In this section, we conduct experiments to answer the following research questions: (\textbf{RQ1}) How effectively and efficiently does \method solve tasks compared to existing search methods? (\textbf{RQ2}) How efficient is \method when conducting a rigorously complete search? (\textbf{RQ3}) Do LLMs inherently possess the capability for self-search?
\begin{table*}[t]
\caption{Results of different search methods across three tasks. Specifically, the results of LLM-based search baselines are reported as the average values across three LLMs. "TL" indicates that the number is too large. We also highlight the SS results of \method that are comparable to state-of-the-art performance (marked in \colorbox{green!36}{green}). 
}
\vspace*{-0.5em}
\centering
\small
\tabcolsep 3.5pt
\renewcommand\arraystretch{1.0}
\begin{tabular}{llccccccc}
\toprule
\multirow{2}{*}{} &
\multirow{2}{*}{Search Method} &
\multicolumn{2}{c}{Game of 24} &
\multicolumn{2}{c}{Mini Crosswords} &
\multicolumn{2}{c}{Blocksworld} \\
\cmidrule(r){3-4} \cmidrule(r){5-6} \cmidrule(r){7-8}
&  & PR (\%) & Avg. SS & PR (\%) & Avg. SS & PR (\%) & Avg. SS \\
\midrule
\multicolumn{8}{l}{\textbf{Traditional Search}} \\
\cmidrule(r){1-2}
&\textsc{Exhaustive Search}&\num{100} &\num{12928} &\num{100} & TL &\num{100} & TL\\
\addlinespace[0.1em]\hdashline\addlinespace[0.1em]
& \textsc{Brute-force (DFS)} & \num{100} & \num{1623} & \num{100} & \num{4128.9} & \num{100} & \num{18531.9} \\

& \textsc{Brute-force (BFS)} & \bfseries\num{100} & \num{3429} & \num{100} & {TL} & \num{100} & \num{96759.4} \\
\midrule
\multicolumn{8}{l}{\textbf{LLM-based Search}} \\
\cmidrule(r){1-2}
& \textsc{Vanilla CoT} & \num{14.3} & 1 ({$\downarrow$} 99.9\%) & \num{1.7} &  1 ({$\downarrow$} 99.9\%) & \num{17.5} &  1 ({$\downarrow$} 99.9\%) \\

& \textsc{Majority Vote} & \num{20.3} & \num{10} ({$\downarrow$} 99.3\%) & \num{3.4} & \num{10} ({$\downarrow$} 99.7\%) & {\num{47.4}} & \num{10} ({$\downarrow$} 99.9\%) \\

& \textsc{Best-of-N} & \num{22.6} & \num{20} ({$\downarrow$} 98.7\%) & \num{2.5} & \num{20} ({$\downarrow$} 99.5\%) & \num{30.0} & \num{20} ({$\downarrow$} 99.8\%) \\

& \textsc{Beam Search} & \num{54.0} & \num{94.8} ({$\downarrow$} 94.1\%) & {\num{45.0}} & \num{26.1} ({$\downarrow$} 99.4\%) & \num{0} & \num{64.4} ({$\downarrow$} 99.6\%) \\
& \textsc{Beam Search}+RV & \num{79.0} & \num{98.8} ({$\downarrow$} 93.9\%) & {\num{90.0}} & \num{30.9} ({$\downarrow$} 99.2\%) & \num{0} & \num{94.8} ({$\downarrow$} 99.6\%) \\
\addlinespace[0.1em]\hdashline\addlinespace[0.1em]
% \rowcolor{gray!30} 
& \textsc{\method} & & & & & & \\

& \;\; -- \textsc{GPT-4o-mini} & $\mathbf{100}$ & \cellcolor{green!36}\num{40} ({$\downarrow$} 97.5\%) & $\mathbf{100}$ & \cellcolor{green!36}\num{75.8} ({$\downarrow$} 98.2\%) & $\mathbf{100}$ & \cellcolor{green!36}\num{160.6} ({$\downarrow$} 99.1\%)\\

% & \;\; -- \textsc{GPT-4o-mini (No Verifier)} & $\mathbf{90}$ & \cellcolor{green!36}\num{44.9} ({$\downarrow$} \%) & $\mathbf{100}$ & \cellcolor{green!36}\num{42.9} ({$\downarrow$} 98.9\%) & $\mathbf{}$ & \cellcolor{green!36}\num{} ({$\downarrow$} \%)\\

& \;\; -- \textsc{GPT-4o} & ${99}$ & \cellcolor{green!36}\num{65.6} ({$\downarrow$} 96.0\%) & $\mathbf{100}$ & \cellcolor{green!36}\num{45.0} ({$\downarrow$} 98.9\%) & $\mathbf{100}$ & \cellcolor{green!36}\num{80.8} ({$\downarrow$} 99.5\%) \\

& \;\; -- \textsc{Qwen2.5-72B-Instruct} & $\mathbf{100}$ & \cellcolor{green!36}\num{84.9} ({$\downarrow$} 94.8\%) & $\mathbf{100}$ & \cellcolor{green!36}\num{98.3} ({$\downarrow$} 98.9\%) & $\mathbf{100}$ & \cellcolor{green!36}\num{68.7} ({$\downarrow$} 99.6\%) \\

\bottomrule
\end{tabular}
% \vspace{-5pt}
\vspace{-1em}
\label{tab:main_result}
\end{table*}

\subsection{Experiment Settings}
\label{sec:experimental_setup}
\noindent \textbf{Tasks}. To evaluate the effectiveness of \method, in addition to the Game of 24 used in Sec.~\ref{sec:preliminary_setup}, we select two widely adopted planning tasks for evaluating LLMs' planning abilities: Mini Crosswords~\cite{yao2023tree} and Blocksworld~\cite{valmeekam2022large}. Further details on these tasks are provided in Appendix~\ref{appendix:tasks}.

% We evaluate the effectiveness of \method across three challenging real-world planning domains: \textbf{(i)} Game of 24~\cite{yao2023tree}; \textbf{(ii)} Mini Crosswords~\cite{yao2023tree}; and \textbf{(iii)} Blocksworld~\cite{valmeekam2022large}. More details of these tasks are in Appendix~\ref{appendix:tasks}.

% \noindent \textbf{Tasks}. We evaluate the effectiveness of \method across three challenging real-world planning domains: \textbf{(i)} Game of 24~\cite{yao2023tree}, which is used in Sec.~\ref{sec:preliminary_setup}; \textbf{(ii)} Mini Crosswords~\cite{yao2023tree}, which involves solving $5 \times 5$ crossword puzzles by selecting words from fixed-length lists corresponding to 5 horizontal and 5 vertical clues; and \textbf{(iii)} Blocksworld~\cite{valmeekam2022large}, a planning task requiring the agent to rearrange blocks into specified stack configurations. More details of these tasks are in Appendix~\ref{appendix:tasks}.

% \noindent \textbf{Baselines}. We compare \method with three LLM-based search algorithms: \textsc{Majority Vote}~\cite{wang2023selfconsistency}, \textsc{Best-of-N}~\cite{snell2024scaling}, and \textsc{Beam Search}~\cite{feng2023alphazerolike}. We also include \textsc{Vanilla CoT}~\cite{wei2022chain} as a reference baseline. For traditional search, we include two \textsc{brute-force} algorithms: depth-first search (\textsc{DFS}) and breadth-first search (\textsc{BFS}). The \textsc{exhaustive search} algorithm is included as a baseline for evaluating search efficiency. Details on the baselines are provided in Appendix~\ref{appendix:baselines}.

\noindent \textbf{Baselines}. Following the settings in Sec.~\ref{sec:preliminary_setup}, we consider four LLM-based searches, \textsc{Vanilla CoT}, \textsc{Majority Vote}, \textsc{Best-of-N}, and \textsc{Beam Search}, and three traditional searches, \textsc{DFS}, \textsc{BFS} and \textsc{Exhaustive Search}. We also involve a baseline \textsc{Beam Search+RV} that adds a rule-based verifier to the beam search method.
Details on the baselines are provided in Appendix~\ref{appendix:baselines}.

\noindent \textbf{Models}. Our experiments use both closed-source models (GPT-4o-mini, GPT-4o~\cite{gpt4ocard}) and open-source models (Qwen2.5-72B-Instruct~\cite{qwen2.5}, QwQ-32B-Preview~\cite{qwq-32b-preview}, DeepSeek-R1~\cite{deepseekai2025deepseekr1}). Note that GPT-4o-mini is a small language model (SLM), and QwQ-32B-Preview and DeepSeek-R1 are the state-of-the-art LRMs. Model and implementation details are in Appendix~\ref{appendix:llm_setting} and~\ref{appendix:implementation_details}.

\noindent \textbf{Evaluation Metrics}.  We follow Sec.~\ref{sec:preliminary_setup} to use two metrics: (i) pass rates (PR) and (ii) search steps (SS).  More details are in Appendix~\ref{appendix:evaluation_metrics}. 

% We evaluate performance using two primary metrics: (i) pass rates (PR) across all problems in the task, measuring solution quality, and (ii) search steps (SS), measuring exploration efficiency by counting traversed states in the search space $\mathcal{S}$.
% More definitions are in Appendix~\ref{appendix:evaluation_metrics}.  

% \noindent \textbf{Evaluation Metrics}.  We evaluate performance using two primary metrics: (i) pass rates (PR) across all problems in the task, measuring solution quality, and (ii) search steps (SS), measuring exploration efficiency by counting traversed states in the search space $\mathcal{S}$.
% More definitions are in Appendix~\ref{appendix:evaluation_metrics}.  

% \noindent \textbf{Implementation Details}.
% Following the settings in prior studies~\cite{yao2023tree, hao2023reasoning, snell2024scaling}, we set the temperature for LLMs to $\num{0.7}$. To use GPT-4o~\cite{gpt4ocard} and GPT-4o-mini~\cite{achiam2023gpt}, we set the mode as the Chat Completion modes for them. For DeepSeek-R1~\cite{deepseekai2025deepseekr1}, we use the Chat mode to run experiments. More details of the implementation are in Appendix~\ref{appendix:implementation_details}.

\subsection{RQ1: Effectiveness and Efficiency Evaluations}
\label{sec:rq1}
% To answer \textbf{RQ1}, we conduct experiments on the three tasks using three different LLM backbones: GPT-4o-mini, GPT-4o, and Qwen2.5-72B-Instruct. For Game of 24, we follow~\cite{yao2023tree} to select $100$ problems indexed from $900$ to $999$. For Mini Crosswords, we randomly choose $20$ problems with $11$ candidate words for each clue, while for Blocksworld, we randomly select $20$ problems with a minimum solution length of $8$ steps. The results are reported in Table~\ref{tab:main_result}.
We conduct experiments on the three tasks using three different LLM backbones: GPT-4o-mini, GPT-4o, and Qwen2.5-72B-Instruct. Detailed task setups are in Appendix~\ref{appendix:rq1_appendix}. The results are reported in Table~\ref{tab:main_result}. The key observations are:
\textbf{(i)} \method significantly reduces search steps compared to brute-force searches. Specifically, it reduces search steps by up to $99.1\%$ compared to \textsc{DFS} and still achieves state-of-the-art search steps compared to other LLM-based methods. This validates \method's efficiency in navigating the search space.
\textbf{(ii)} \method achieves a near-perfect pass rate across all settings, outperforming other LLM-based methods.
% , which achieve pass rates of up to 54.0\%, 45.0\%, and 47.4\% for the three tasks, respectively. This validates the effectiveness of \method in problem-solving.
\textsc{Beam Search+RV} achieves better pass rates than \textsc{Beam Search}, indicating that integrating rule-based verifiers into LLM-based methods enhances their reliability in problem-solving. However, it still falls behind \method, confirming that current LLM-based methods do not fully conduct real search. This further highlights the superior search effectiveness of \method.
Full search results and the discussion of \method with various verifiers are in Appendix~\ref{appendix:additional_results_seal}.
% Appendix~\ref{appendix:full_comparison_results_on_three_tasks_three_LLMs} and~\ref{appendix:impact_verifier_search}.

% $2.8\times 10^{11}$ $5.3\times 10^{9}$

\begin{figure*}[t]
    \small
    \centering
    \begin{subfigure}{0.24\textwidth}
        \includegraphics[width=0.98\linewidth]{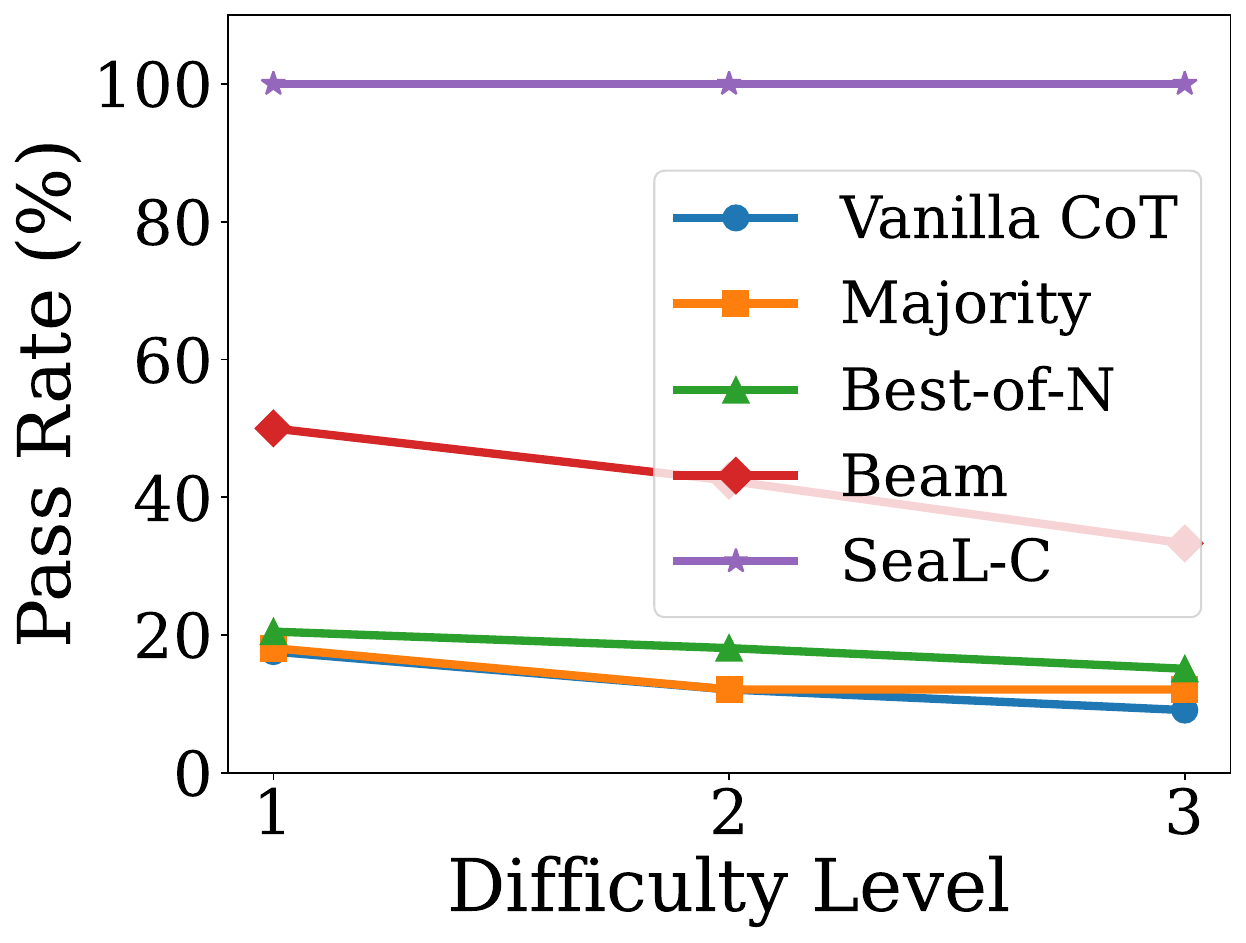}
        \vskip -0.5em
        \caption{Game of 24 - PR}
    \end{subfigure}
    \begin{subfigure}{0.24\textwidth}
        \includegraphics[width=0.98\linewidth]{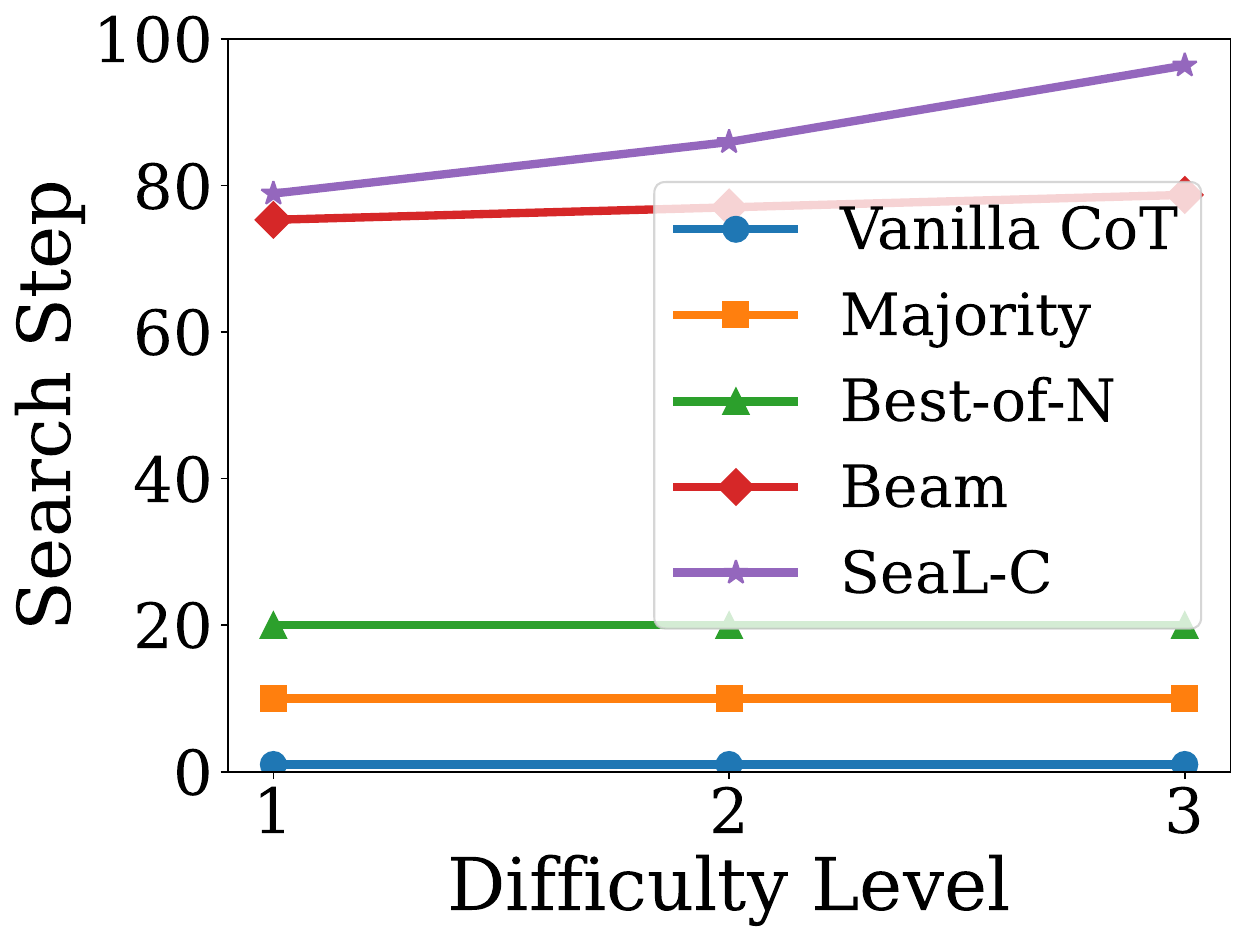}
        \vskip -0.5em
        \caption{Game of 24 - SS}
    \end{subfigure}
    \begin{subfigure}{0.24\textwidth}
        \includegraphics[width=0.98\linewidth]{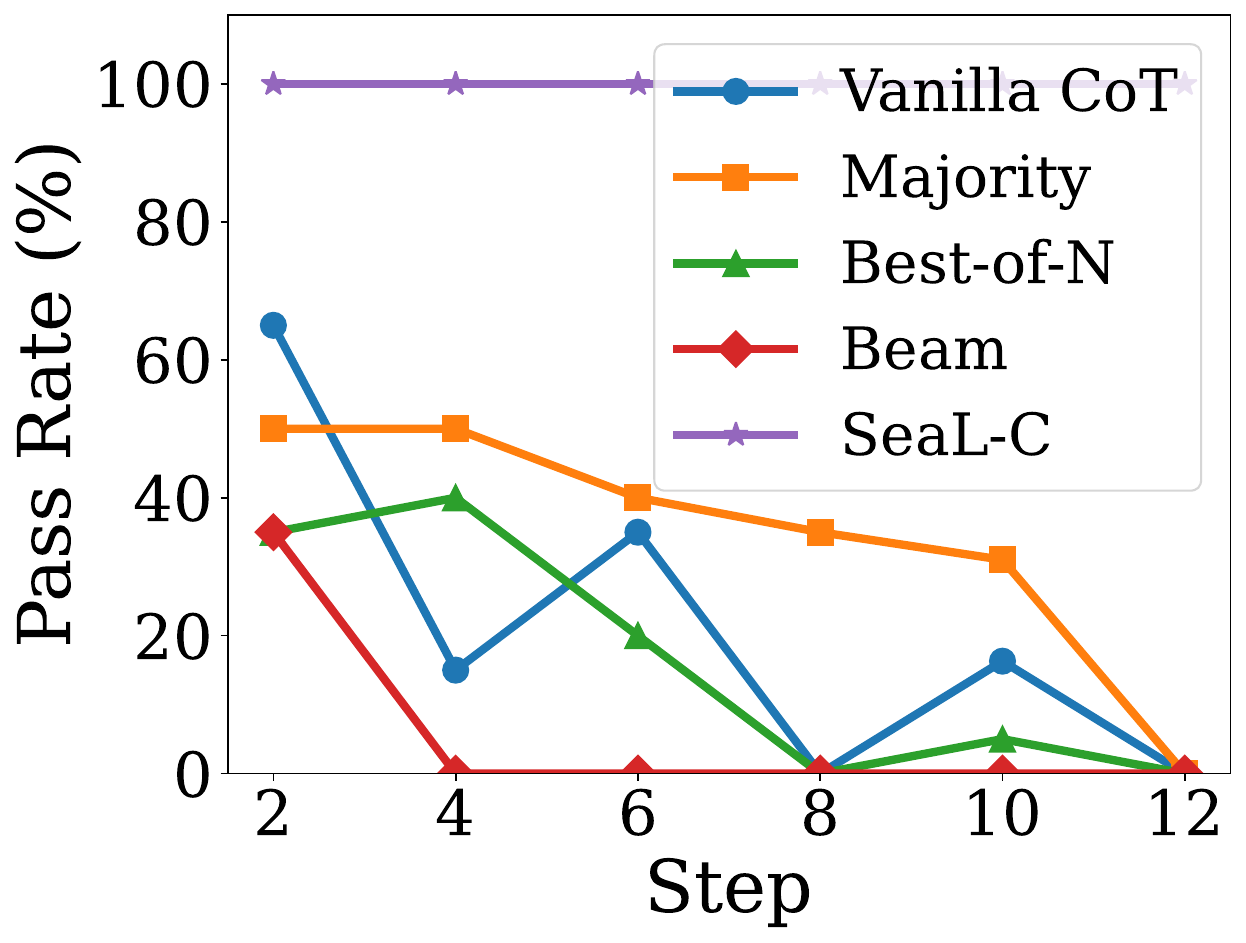}
        \vskip -0.5em
        \caption{Blocksworld - PR}
    \end{subfigure}
    \begin{subfigure}{0.24\textwidth}
        \includegraphics[width=0.98\linewidth]{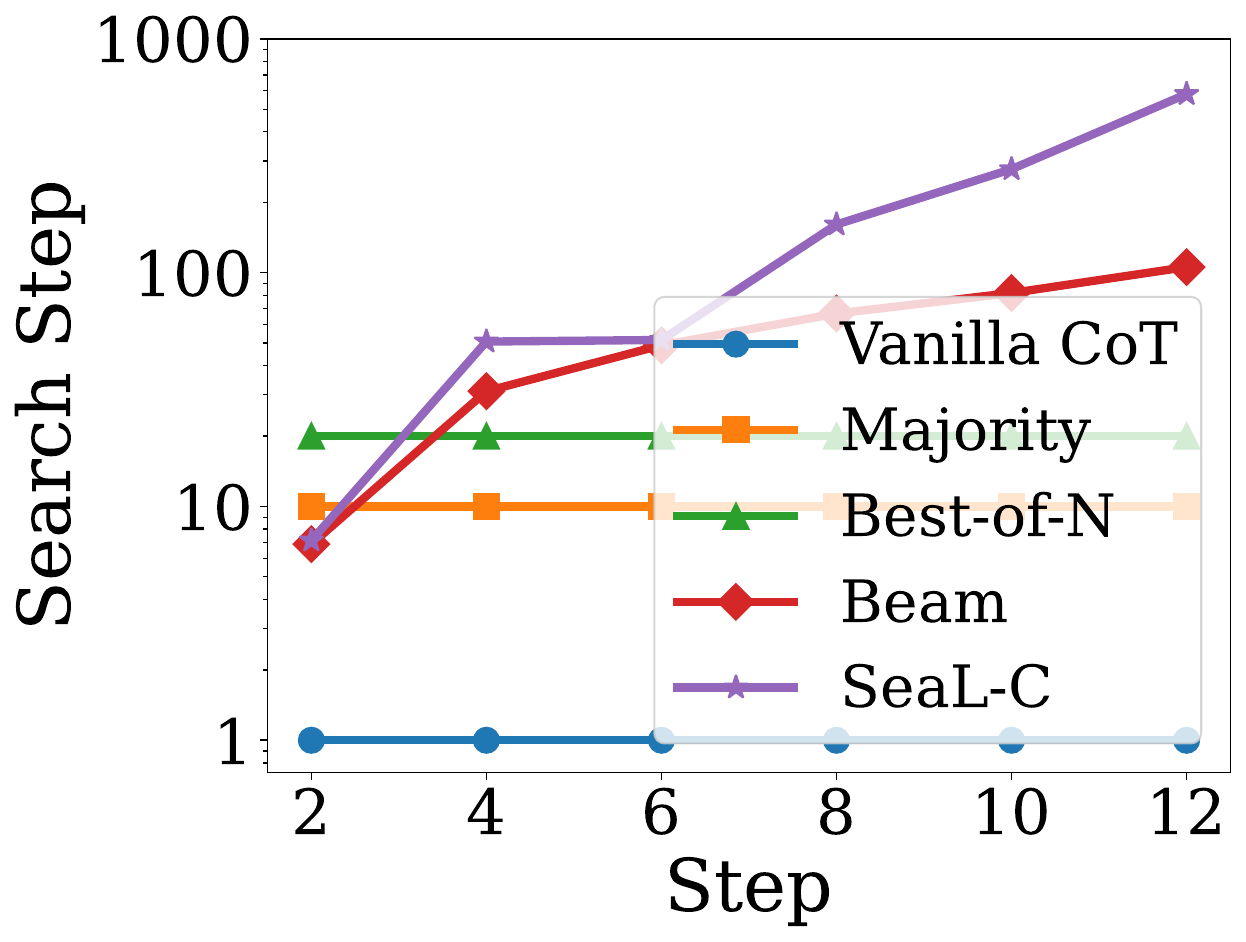}
        \vskip -0.5em
        \caption{Blocksworld - SS}
    \end{subfigure}
    \vskip -1.em
    \caption{Impact of problem difficulty on search completeness in Game of 24 and Blocksworld with GPT-4o-Mini.}
    % \vskip -2em
    \label{fig:impact_problem_difficulty_completeness}
\end{figure*}

\subsection{RQ2: Impact of Problem Difficulty on Search Completeness}
\label{sec:rq2}
% \minhua{revise to select two tasks}
We evaluate \cmethod's performance across problem difficulties in three tasks. For Game of 24, we use three difficulty levels (Sec.~\ref{sec:preliminary_results_analysis}); for Blocksworld, difficulty scales with minimum required action steps ($2$–$12$). Other settings follow Sec.~\ref{sec:experimental_setup}. The results using GPT-4o-mini in the two tasks are reported in Fig.~\ref{fig:impact_problem_difficulty_completeness}, which show that: \textbf{(i)} \cmethod achieves $100\%$ pass rates across all difficulty levels, outperforming baseline methods whose performance degrades progressively, confirming its completeness. \textbf{(ii)} As the difficulty increases, \cmethod's search steps increase slightly but remain significantly lower than brute-force baselines. This efficiency is attributed to \cmethod's learning-guided complete state decomposition and two-phase ranking, which prioritize promising states and effectively reduce the search space. Mini Crosswords results are in Appendix~\ref{appendix:addtional_impact_difficulty_of_seal_c}.
% We explore \cmethod's performance across varying problem difficulties in three tasks. For Game of 24, problems are categorized into three difficulty levels as in Sec.~\ref{sec:preliminary_results_analysis}.
% For blocksworld, difficulty varies by the minimum action steps required, ranging from $\{2, 4, 6, 8, 10, 12\}$. Other settings follow Sec.~\ref{sec:experimental_setup}. The results using GPT-4o-mini in Game of 24 and Blocksworld tasks are shown in Fig.~\ref{fig:impact_problem_difficulty_completeness}. From the figures, we observe: \textbf{(i)} \cmethod consistently maintain a $100\%$ pass rate across all difficulty levels in the two tasks, while the pass rate of other methods gradually decrease. This satisfies our expectations and validates the completeness of \cmethod. \textbf{(ii)} As the difficulty increases, \cmethod's search steps increase slightly but remain significantly lower than brute-force baselines. This efficiency is attributed to \cmethod's learning-guided complete state decomposition and two-phase ranking, which prioritize promising states and effectively reduce the search space.
% Additional results on Mini Crosswords are in Appendix~\ref{appendix:addtional_impact_difficulty_of_seal_c}.
% For Mini Crosswords, we vary the length of the candidate word lists as $\{6,7,8,9,10,11\}$ to define difficulty levels.

\subsection{RQ3: Potential of LLMs in Self-Search}
\label{sec:experiment_self_search}
We conduct an initial exploration to study whether LLMs can learn to self-search for problem-solving by testing two self-search prompts in Sec.~\ref{sec:llm_self_search} and $\textsc{Vanilla CoT}$ in GPT-4o and two state-of-the-art open-sourced LRMs, QwQ-32B-Preview and DeepSeek-R1 to demonstrate the importance of search in solving tasks. Especially, we use the chat mode of DeepSeek-R1 and sample $20$ problems.
Since they are LRMs that are capable of searching, the results of \textsc{Self-Search (High)} in the two LRMs can be directly obtained from using $\textsc{Vanilla CoT}$ in this model.
The comparison results in Table~\ref{tab:self_search_result} show: 
\textbf{(i)} Standard prompting (\textsc{Vanilla CoT}) in GPT-4o achieves the lowest performance, indicating LLMs cannot conduct searches and require conduct searches for complex problem-solving.
{\textbf{(ii)} Two LRMs achieve the best performance across all settings. Specifically, DeepSeek-R1, when using \textsc{Self-Search (High)}, achieves an impressive pass rate of \num{85}\%. From our observations, these LRMs perform search by iteratively sampling and evaluating different answers until a correct solution is found. This underscores the importance of search in enhancing LLM problem-solving capabilities while also revealing the inefficiency of current search mechanisms, motivating the need for more efficient strategies. More details of the problem-solving processes of the two LRMs are in Appendix~\ref{appendix:more_details_of_problem_solving_processes_LRMs}.}
\textbf{(iii)} Both self-search approaches significantly improve performance, with low-level self-search achieving up to a $95\%$ pass rate. This demonstrates that LLMs have the potential of effectively utilizing explicit search strategies for improving problem-solving.
\textbf{(iv)} Low-level self-search, which incorporates \method's strategies, consistently outperforms high-level self-search across both models. This validates the effectiveness of \method's search and suggests that better search guidance enhances LLM reasoning. 

These findings highlight both the importance of search in enhancing LLM capabilities and the effectiveness of \method's search strategies, motivating us to explore how to further develop LLMs' self-search capabilities and optimize search strategies, which will be our future works. 

\begin{table}[t]
\caption{PR(\%) of search strategies for GPT-4o, QwQ-32B-Preview and DeepSeek-R1 on the Game of 24.}
% \vspace{-0.5em}
\centering
\small
\tabcolsep 3.5pt
\renewcommand\arraystretch{1.0}
\begin{tabular}{llc}
\toprule
\multirow{1}{*}{Model} &
\multirow{1}{*}{Method} &
PR(\%) \\
\midrule
\multirow{3}{*}{GPT-4o} &
\textsc{Vanilla CoT} & \num{13} \\
& \textsc{Self-search (High)}  & \num{24} \\
& \textsc{Self-search (Low)}  & $\mathbf{31}$ \\
\midrule
\multirow{2}{*}{QwQ-32B-Preview} &
\textsc{Self-search (High)}  & \num{62} \\
& \textsc{Self-Search (Low)}  & $\mathbf{70}$ \\
\midrule
\multirow{2}{*}{DeepSeek-R1 (20)} &
\textsc{Self-search (High)}  & \num{85} \\
& \textsc{Self-Search (Low)}  & $\mathbf{95}$ \\
\bottomrule
\end{tabular}
% \vspace{-2em}
\label{tab:self_search_result}
\end{table}

\subsection{Ablation Studies}
\begin{figure}[t]
    \small
    \centering

    \includegraphics[width=0.55\linewidth]{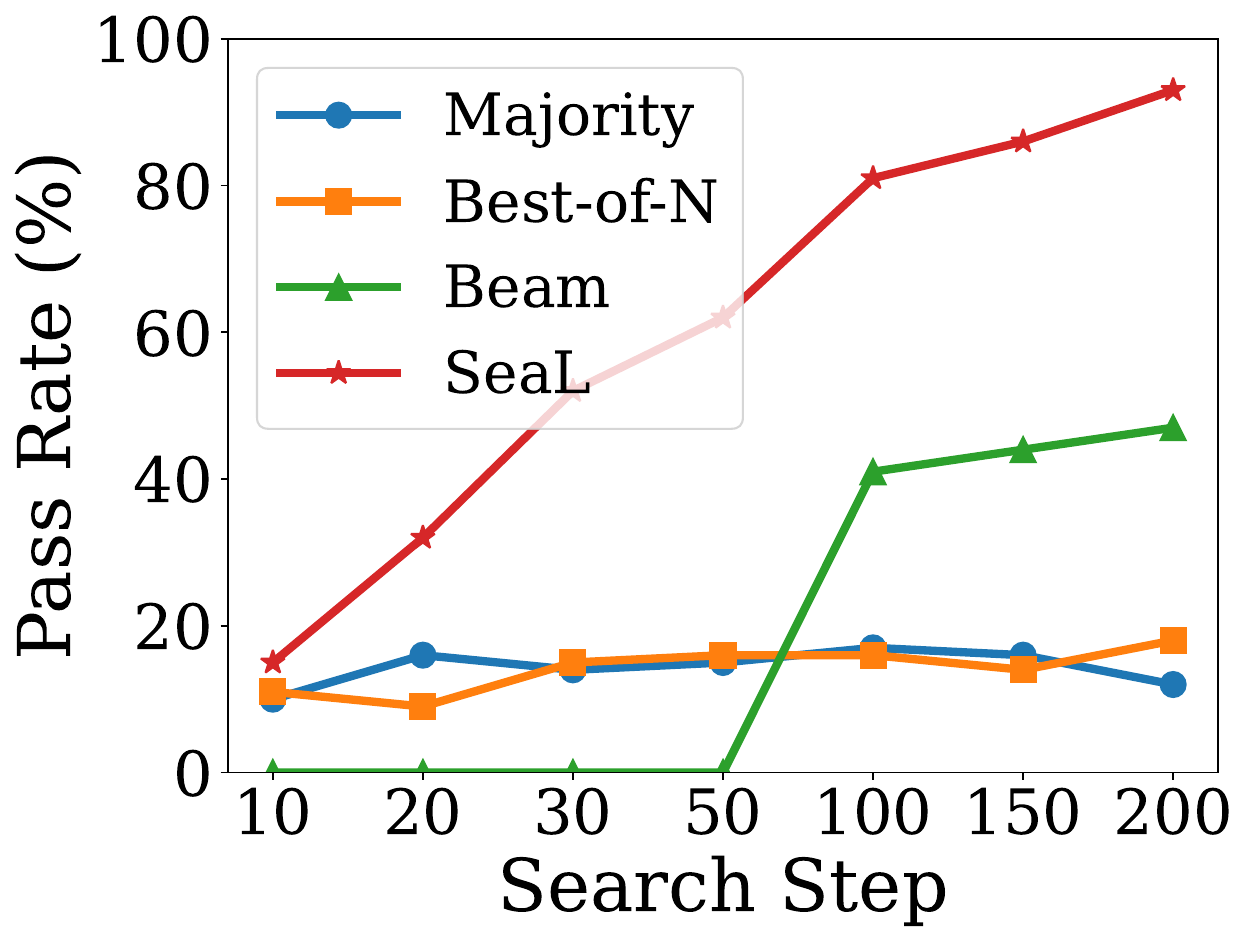}
    \vskip -1em
    \caption{Impact of SS in \method using GPT-4o-mini.}
    \vskip -2em
    \label{fig:impact_of_search_number_budget_ablation}
\end{figure}

Inspired by \citet{snell2024scaling}, we conduct ablation studies to understand the impact of search budgets on \method's performance. Instead of terminating the search upon finding the final state, we introduced a pre-defined search step budget, where \method terminates early if the budget is reached. We vary the search step budgets as $\{10, 20, 30, 50, 100, 150, 200\}$, and compare various search methods using GPT-4o-mini in Game of 24. The comparison results are reported in Fig.~\ref{fig:impact_of_search_number_budget_ablation}. Our analysis reveals three key findings: \textbf{(i)} \method consistently outperforms other methods across all search budgets, demonstrating its effectiveness in accurately solving problems even under constrained search budgets. \textbf{(ii)} Pass rates for all search methods generally improve as the search budget increases, aligning with expectations that scaling test-time computation enhances search performance. 
\textbf{(iii)} When the search budget is small (less than $50$ steps), \textsc{Beam Search} performs poorly, achieving a 
$0\%$ pass rate. This is due to its sequential evaluation of substates, which consumes substantial search budgets on generating and assessing intermediate substates. In contrast, \method focuses on diving directly toward the goal state, significantly reducing the search space and validating its efficiency in navigating complex search problems. Additional results from the ablation studies evaluating the impact of \method's components can be found in Appendix~\ref{appendix:additional_results_of_ablation_studies}.

\section{Conclusion and Future Work}
In this paper, inspired by the principles of search and learning~\cite{Sutton2019BitterLesson,snell2024scaling}, we systematically investigate the integration of learning into search. We first explore how learning benefits the search process via LLMs, demonstrating that LLMs improve search efficiency by reducing search space.
Then, building on these insights, we introduce a novel framework, \method, and its variant \cmethod, designed to combine the reasoning capabilities of LLMs with search strategies to achieve efficient and accurate problem-solving. 
Extensive experiments conducted on three real-world planning tasks demonstrate that \method achieves near-perfect pass rates across various settings while significantly reducing search spaces, showcasing the effectiveness and efficiency of \method. Furthermore, we also explore how search can benefit LLMs, evaluating whether LLMs can develop self-search capabilities. 
We show that search significantly enhances their reasoning and learning performance. These findings highlight the bidirectional synergy between search and learning, emphasizing the potential of integrating search into LLMs.

Our research 
% bridges the gap between systematic search and learning-driven reasoning, 
paves the way for further exploration into the convergence of search and learning. Future work will focus on investigating how LLMs can better conduct self-searches, further unlocking their potential for complex problem-solving.

\section{Limitations}
One potential limitation of this work is the necessity of encoding our search strategies into prompts to enable LLMs to perform self-search. In the future, it is worth exploring how to allow LLMs to autonomously conduct self-search during reasoning without explicit supervision within the prompts. Additionally, our work primarily focuses on LLMs; an important direction for future investigation is to assess the applicability of our search strategy to multi-modal LLMs.

\section{Impact Statements}
This paper introduces \method and \cmethod, frameworks designed to enhance search processes by integrating the reasoning capabilities of LLMs with structured search strategies. 
% Our work aims to advance the field of AI by addressing the challenges of efficiency and completeness in search algorithms, enabling LLMs to solve complex problems more effectively.
The potential societal impacts of this research are broad but largely align with the established consequences of improving computational problem-solving techniques. Enhanced search efficiency and the ability to leverage LLMs for self-search could benefit applications in diverse fields such as healthcare, logistics, education, and robotics, where intelligent decision-making is crucial. 
However, as with all advancements in AI, there are ethical considerations. For example, the misuse of improved search strategies in domains such as automated surveillance or adversarial systems could lead to privacy or security concerns. We encourage researchers and practitioners to apply this work responsibly and ensure it aligns with ethical guidelines.
We see no immediate risks or unintended negative consequences specific to this work that require urgent attention. This paper primarily contributes to foundational research in search and learning integration. Future exploration of self-search capabilities in LLMs will include careful assessment of ethical implications to ensure responsible development and deployment of these technologies.

% In the unusual situation where you want a paper to appear in the
% references without citing it in the main text, use \nocite
\nocite{langley00}

\bibliography{ref}
\clearpage
\appendix
% \newpage
\section*{Appendix Overview}
\begin{itemize}[leftmargin=*]
\item \textbf{Appendix~\ref{appendix:related_works}}: Related Work.
\item \textbf{Appendix~\ref{appendix:section_title_more_details_of_preliminary_analysis}}: More details of Preliminary Analysis.
\item \textbf{Appendix~\ref{appendix:algorithm_full}}: Algorithm.
\item \textbf{Appendix~\ref{appendix:additional_details_seal_sealc}}: Additional Details of \method and \cmethod.
\item \textbf{Appendix~\ref{appendix:more_details_self_search}}: Additional Details of Self-Search.
\item \textbf{Appendix~\ref{appendix:experimental_settings}}: Experimental Settings.
\item \textbf{Appendix~\ref{appendix:rq1_appendix}}: Full Analysis of RQ1: Effectiveness and Efficiency Evaluations.

\item \textbf{Appendix~\ref{appendix:additional_results_seal}}: Additional Experimental Results of \method.
\item \textbf{Appendix~\ref{appendix:addtional_impact_difficulty_of_seal_c}}: Additional Experimental Results of \cmethod.
\item \textbf{Appendix~\ref{appendix:more_details_of_problem_solving_processes_LRMs}}: Additional Details of Problem-solving Processes of LRMs.
\item \textbf{Appendix~\ref{appendix:additional_results_of_ablation_studies}}: Additional Experimental Results of Ablation Studies.
\end{itemize}

\section{Full Details of Related Works}
\label{appendix:related_works}
\subsection{LLM-based Search Methods}
\label{appendix:related_works_LLM_search}
Recent advancements in test-time compute scaling have sparked growing interest in methods that enable LLMs to simulate search processes and ``think longer'' instead of directly generating answers in one pass. Several works~\cite{wang2023selfconsistency,hao2023reasoning,feng2023alphazerolike,yao2023tree,zhao2024large,besta2024got,wang2024litesearch,snell2024scaling} adopt such approaches to enhance LLMs' problem-solving capabilities. Self-Consistency~\cite{wang2023selfconsistency} employs majority voting over multiple sampled reasoning chains, based on the intuition that the solutions for complex problems are rarely unique. Similarly, Best-of-N~\cite{snell2024scaling} samples multiple outputs and selects the highest-scoring solution using a learned verifier or reward model~\cite{cobbe2021training,lightman2023let}. 
RAP~\cite{hao2023reasoning} uses LLMs as heuristic policy functions to guide Monte Carlo Tree Search (MCTS), treating LLMs as world models for state exploration. Tree of Thoughts (ToT)~\cite{yao2023tree} simulates a search tree with LLM-guided node expansion and pruning, while Graph of Thoughts (GoT)~\cite{besta2024got} generalizes this approach by modeling the search space as a graph. Despite their innovation, these methods rely solely on LLMs' intrinsic knowledge, often leading to unstable performance due to limitations in LLMs' reasoning capabilities. In contrast, our proposed \method integrates LLMs with traditional search strategies, ensuring both completeness and efficiency in solving decision-making tasks.

\subsection{Traditional Search Methods}
\label{appendix:related_works_traditional_search}
Traditional search methods have been extensively utilized in tasks such as combinatorial optimization~\cite{crama2005local} and pathfinding~\cite{hart1968formal}. Among these, brute-force search is regarded as a general yet powerful approach~\cite{Sutton2019BitterLesson}, offering guaranteed solutions through exhaustive exploration of the search space. However, this method also suffers from the large search space in complex problems.  
Inspired by the recent discussion about search and learning~\cite{Sutton2019BitterLesson}, which underscores the enduring value of general-purpose strategies that scale with computational power, we investigate how to better integrate search with learning to leverage the strengths of both paradigms. Our \method and \cmethod draw inspiration from traditional search while incorporating LLM-guided reasoning to reduce search space, thereby significantly enhancing efficiency without sacrificing completeness.

\subsection{LLMs for Planning and Decision-Making}  
Planning and decision-making involve devising strategic action sequences to achieve predefined goals from given initial states. While classical planning problems rely on algorithms such as brute-force search and A*~\cite{hart1968formal} for optimal plan generation, recent advancements demonstrate that LLMs can leverage their extensive commonsense knowledge for planning~\cite{valmeekam2023planning,xiao2023llm,yao2023react}. 
% Several works use LLMs as heuristic policy functions for generating feasible plans. For example, \citeauthor{valmeekam2023planning}~(\citeyear{valmeekam2023planning}) analyzed LLMs' capabilities in classical planning tasks. 
For example, LLM A*~\cite{xiao2023llm} integrates LLMs into the A* algorithm by using them to generate intermediate waypoints, improving efficiency in robotics and navigation tasks. Other works~\cite{lyu2023faithful,jojic2023gpt,liu2023llm+,katz2024planning,cao2024automating} exploit LLMs' programming abilities for planning tasks. For instance, LLM+P~\cite{liu2023llm+} translates task instructions into Planning Domain Definition Language (PDDL) and solves them using classical algorithms like A*. Thoughts of Search (ToS)~\cite{katz2024planning} uses LLMs to generate planning code with human feedback, while AutoToS~\cite{cao2024automating} automates this loop by enabling LLMs to validate and revise their own code.

\subsection{Reasoning with LLMs}  
Chain-of-Thoughts (CoT)~\cite{wei2022chain} was the first major breakthrough revealing that LLMs can formulate multi-step reasoning processes by using explicit prompts like ``Let's think step by step.'' Follow-up works~\cite{wang2023selfconsistency,yao2023react,zhou2023leasttomost,welleck2023generating,shinn2024reflexion,paul2023refiner} further build on this paradigm. ReAct~\cite{yao2023react} integrates reasoning with planning, interleaving reasoning steps with dynamic interactions. PAL~\cite{gao2023pal} enhances CoT by leveraging LLMs' programming abilities, guiding them to generate executable code during reasoning. 
Several recent approaches~\cite{welleck2023generating,shinn2024reflexion,paul2023refiner} introduce self-evaluation capabilities, enabling LLMs to provide feedback on their intermediate reasoning steps. 
Research has also expanded across domains~\cite{wang2024comprehensive,zhang2024divide,zhang2024does,xu2024llm}, including time series analysis~\cite{lin2024decoding} and vision-language integration~\cite{wu2025lanp}
In this work, we systematically study how LLMs' reasoning capabilities can complement traditional search methods, enabling accurate and efficient problem-solving through our proposed \method framework. 

\section{More details of Preliminary Analysis}
\label{appendix:section_title_more_details_of_preliminary_analysis}

\subsection{Full Analysis: How Learning can Benefit Search}
\label{appendix:preliminary_results_analysis_appendix}
The analysis in Sec.\ref{sec:analysis_existing_search} highlights the limitations of existing search methods and underscores the need for a more effective search-and-learning framework. To better understand how learning can enhance search, we divide the 100 problems from the Game of 24 into three difficulty levels based on human success rates\footnote{{https://www.4nums.com/game/difficulties/}}. Other experimental settings follow those in Sec.~\ref{sec:preliminary_setup}.
Our experimental results, presented in Fig.~\ref{fig:preliminary_difficulty_level_group}, reveal several key findings regarding the performance of existing search methods. Overall, we observe that \textbf{(i)} the pass rates of LLM-based methods decrease as difficulty increases, and \textbf{(ii)} while LLM-based searches require fewer search steps, they generally achieve lower pass rates compared to traditional approaches.

\noindent \textbf{Obs. 1: LLMs Perform Better for Simpler Problems.}. Our analysis demonstrates that LLM-based methods achieve notably higher pass rates on less complex tasks. As illustrated in Fig.~\ref{fig:preliminary_difficulty_level_group}(a), \textsc{Vanilla CoT} attains pass rates of $\num{18.2}\%$, $\num{12.1}\%$, and $\num{11.7}\%$ for problems with difficulty levels $1$, $2$, and $3$, respectively. This pattern suggests that LLMs excel at direct problem-solving for simpler cases, indicating their potential for single-step solutions rather than requiring iterative approaches. More examples are provided in Fig.~\ref{fig:obs1_example}.

\noindent \textbf{Obs. 2: Learning-Based Pruning Has Precision-Coverage Trade-offs.}. Among these LLM-based search methods, \textsc{Beam Search} achieves superior performance in pass rate. Specifically, according to Fig.~\ref{fig:preliminary_difficulty_level_group}(a), \textsc{Beam Search} achieves up to a $\num{48.5}\%$ pass rate, while other LLM-based methods only achieve up to a $\num{33}\%$ pass rate. In addition, as in Fig.~\ref{fig:preliminary_difficulty_level_group}(b), \textsc{Beam Search} significantly saves $92.4\%$ SS than \textsc{Brute-force (DFS)}. To deeply analyze this observation, we present an example of \textsc{Beam Search} in solving this task in Fig.~\ref{fig:obs2_example}. From this figure, we observe that in each time of generating next intermediate steps, it only relies on LLMs to generate a few of possible next steps instead of decomposing all possible states in brute-force searches, which significantly reduces the search space and improves search efficiency. However, we also observe from Fig.~\ref{fig:obs2_example} that it is possible that some valid states are overlooked during the intermediate step generation, leading to failure in solving tasks and reflecting a trade-off between efficiency and completeness in search.

% \minhua{Plot an example to show the advantage and limitations of learning-based prunning}.

\noindent \textbf{Obs. 3: Learning Provides Adaptive Search Guidance.}. 
As in Fig.~\ref{fig:preliminary_difficulty_level_group}(a), compared with \textsc{Vanilla CoT}, other LLM-based methods perform better. Specifically, \textsc{Vanilla CoT} only achieve up to $18.2\%$ pass rate, while \textsc{Majority Vote}, \textsc{Best-of-N} and \textsc{Beam Search} achieve up to $29.4\%$, $33.3\%$ and $48.5\%$ prateates, respectively. We analyze that this performance gap is because these LLM-based methods spend more search budget in verifying results via LLMs according to Fig.~\ref{fig:preliminary_difficulty_level_group}, demonstrating LLMs' capability as dynamic evaluators to leverage learned knowledge to identify promising states and guide the search process. Fig.~\ref{fig:obs3_example} is an example of \textsc{Beam Search} leveraging this observation, where \textsc{Beam Search} employs an LLM verifier to assess intermediate states' progress toward the goal, prioritizing exploration of valid states and avoiding further exploration on other states. However, we also notice that some valid states are overlooked mistakenly. This implies that further improvements are needed to ensure critical states are not omitted.

These findings suggest that learning can substantially improve search efficiency when properly integrated into the search process. However, they also highlight the need for careful mechanism design to balance the benefits of learning-based pruning with the completeness guarantees of traditional search. This motivates our development of a systematic framework for learning-enhanced search.

\subsection{Illustrative Examples}
\label{appendix:more_details_pa}
\begin{figure*}[t]
    \small
    \centering
    
        \includegraphics[width=0.5\linewidth]{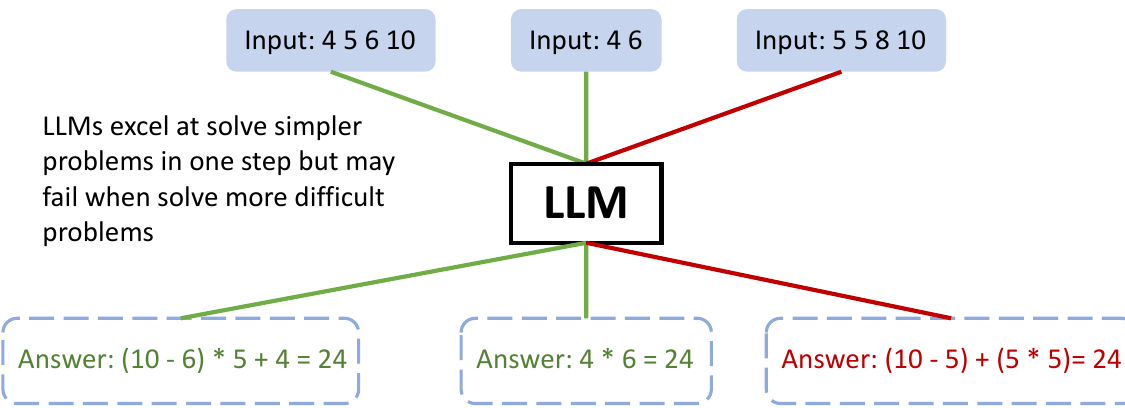}
        \vskip -0.5em

    % \vskip -1.em
    \caption{Illustrative examples of Observation 1 in Sec.~\ref{sec:preliminary_results_analysis}. This figure presents three examples of calculating $24$ from intermediate steps in the Game of 24 task. The \textcolor{green}{green} answer represents a correct equation that results in $24$, whereas the \textcolor{red}{red} answer represents an incorrect equation that does not equal $24$. We observe that LLMs perform well in solving simple tasks in one step, such as $4 \times 6 = 24$ and $(10 - 6) \times 5 + 4 = 24$, but struggle with more complex tasks in a single step (e.g., the third example, where the model fails to find a solution using numbers $5$, $5$, $8$, and $10$).}
    % \vskip -1em
    \label{fig:obs1_example}
\end{figure*}

\begin{figure*}[t]
    \small
    \centering
    
        \includegraphics[width=0.9\linewidth]{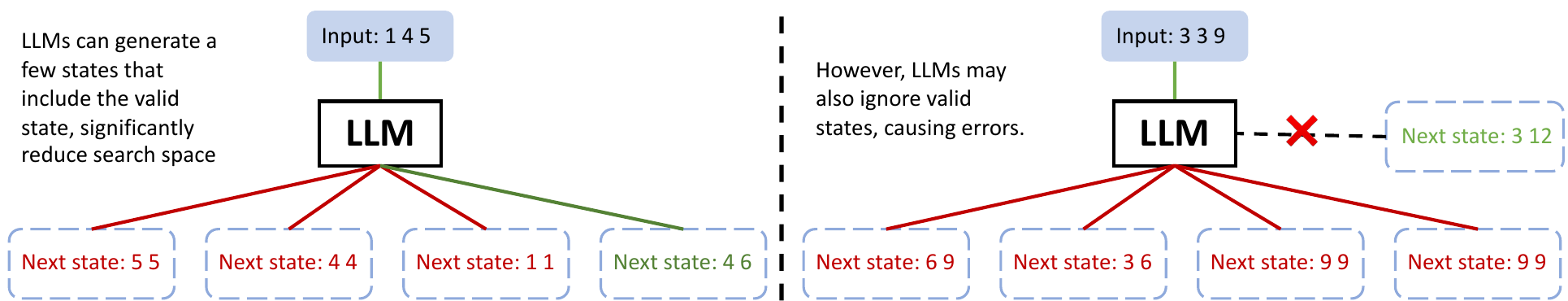}
        \vskip -0.5em

    % \vskip -1.em
    \caption{Illustrative examples of Observation 2 in Sec.~\ref{sec:preliminary_results_analysis}.
This figure presents two examples of using \textsc{Beam Search} for state decomposition to generate the next substates in the game of 24 task. Specifically, the method selects two numbers from the input set and applies a basic arithmetic operation ($+, -, \times, \div$) to produce a new set of numbers.
The \textcolor{green}{green} next state represents an intermediate state that can lead to 24, whereas the \textcolor{red}{red} next state denotes an invalid intermediate state that cannot lead to 24. The left example demonstrates that LLMs can generate a few valid next steps (e.g., selecting $4, 6$ in this case) rather than exhaustively decomposing all possible states through brute-force search. The right example illustrates that LLMs may sometimes overlook valid states and produce only invalid decomposed states.
    }
    % \vskip -1em
    \label{fig:obs2_example}
\end{figure*}

\begin{figure*}[t]
    \small
    \centering
    
        \includegraphics[width=0.9\linewidth]{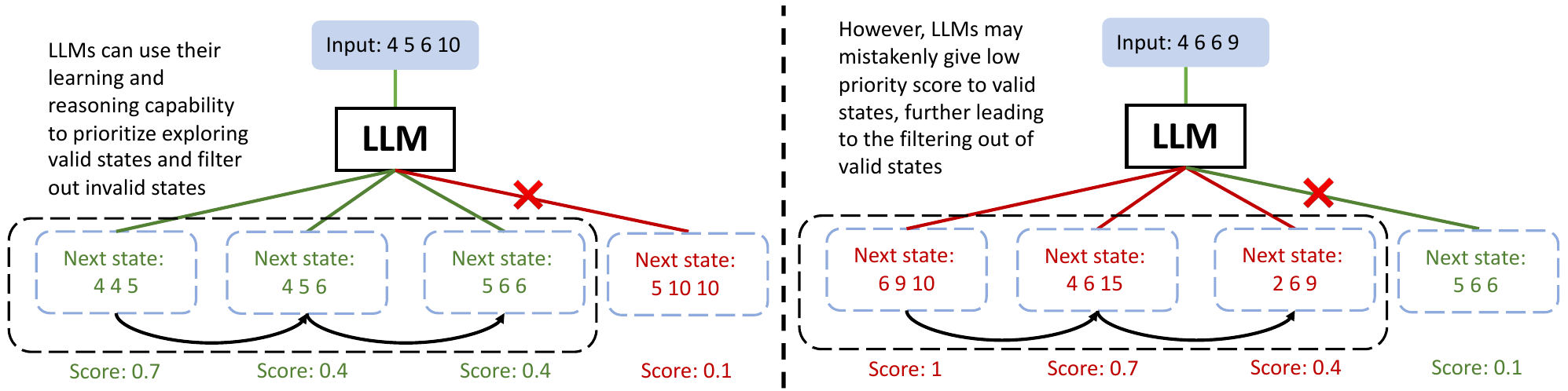}
        \vskip -0.5em

    % \vskip -1.em
    \caption{ Illustrative examples of Observation 3 in Sec.~\ref{sec:preliminary_results_analysis}. This figure presents two examples of using an LLM as a verifier to assess intermediate states' progress toward the goal of 24 in the Game of 24 task. The \textcolor{green}{green} next state represents an intermediate state that can lead to 24, whereas the \textcolor{red}{red} next state denotes an invalid intermediate state that cannot lead to 24.
    In the left example, given the input numbers $4, 5, 6, 10$, the LLM effectively assigns high scores to valid next states (e.g., $4, 4, 5$ with a score of $0.7$, where $4 \times 5 + 4 = 24$) and low scores to invalid next states (e.g., $5,10,10$ with a score of $0.1$, which cannot achieve 24). In contrast, the right example, given the input numbers $4, 6, 6, 9$, demonstrates that LLMs may mistakenly assign high priority scores to invalid states. For instance, $6, 9, 10$ receives the highest score of $1$ but cannot achieve 24. This highlights the need for further improvements to ensure that critical states are not overlooked.}
    % \vskip -1em
    \label{fig:obs3_example}
\end{figure*}

Fig.~\ref{fig:obs1_example},~\ref{fig:obs2_example}, and~\ref{fig:obs3_example} provide the illustrative examples of the observations presented in Sec.~\ref{sec:preliminary_results_analysis}. The full analyses are provided in Appendix~\ref{appendix:preliminary_results_analysis_appendix}. Detailed explanations of these examples can be found in the respective figure captions. Specifically:
\begin{itemize}[leftmargin=*]
    \item Fig.~\ref{fig:obs1_example} illustrates Observation 1 in Sec.~\ref{sec:preliminary_results_analysis}.
    \item Fig.~\ref{fig:obs2_example} presents examples of Observation 2 in Sec.~\ref{sec:preliminary_results_analysis}.
    \item Fig.~\ref{fig:obs3_example} demonstrates Observation 3 in Sec.~\ref{sec:preliminary_results_analysis}.
\end{itemize}

% This demonstrates that LLMs can then use their learning and reasoning capability to priorize exploring valid states and filter out invalid states.
\subsection{Full Analysis: Existing Search Algorithms}
\label{appendix:analysis_existing_search_appendix}
As discussed in Sec.~\ref{sec:introduction}, while existing search algorithms show significant promise, they still encounter notable challenges. Traditional search algorithms, while thorough, suffer from inefficiency due to exhaustive exploration of the state space $\mathcal{S}$. They lack the human-like ability to quickly identify promising paths or recognize dead ends. Conversely, LLM-based search methods, while more efficient, struggle with reliability. They heavily rely on LLMs' reasoning capabilities, which can be unstable or incomplete, particularly in complex scenarios requiring careful exploration of multiple solution paths.

To better illustrate these limitations, we conduct an analysis based on the experimental setup in Sec.\ref{sec:preliminary_setup}, applying existing search algorithms to the Game of 24 task.
The results are reported in Table~\ref{tab:preliminary_24game_qwen}, revealing the following insights: \textbf{(i)} Traditional brute-force algorithms achieve perfect $100\%$ accuracy but require up to \num{3429} search steps, demonstrating their inefficiency in searching for answers.
\textbf{(ii)} While LLM-based methods significantly reduce the search space (at least $92.4\%$ fewer steps than \textsc{Brute-force (DFS)}), their pass rates peak at only $35\%$, highlighting their instability compared to traditional searches.

These findings underscore the necessity of a framework that combines the completeness of systematic search with the efficiency of learning-based approaches. This motivates the development of a more effective search-and-learning methodology to address these limitations.

\section{Algorithm}
\label{appendix:algorithm_full}
% In this section, we provide the implementation algorithm of \method on three tasks.
The algorithm of \method is provided in Alg.~\ref{alg:Framwork}.  This design allows \method{} to systematically integrate learning into the search process, reducing unnecessary state exploration and improving scalability for complex problems.

The algorithm of \cmethod is provided in Alg.~\ref{alg:Framwork_completeness}. This design removes the state validity checking to prevent it from compromising completeness. We use the learning-guided complete state decomposition and two-phase ranking to improve the search efficiency.

\begin{algorithm}[t!] 
\caption{Algorithm of {\method}.}
\label{alg:Framwork} 
\begin{algorithmic}
% \REQUIRE $\mathcal{G}=(\mathcal{V}, \mathcal{E}, \mathbf{X})$, $\mathcal{Y}_L$, $\beta$, $T$.
\REQUIRE Initial state $s^{init}$, goal state $s^{goal}$, Language Model $LLM$, external verifier $f$.
\ENSURE Action sequence $\mathcal{A}$ transforming $s^{init}$ to $s^{goal}$

\FUNCTION{F($s^{cur}$, $M$, $f$, $\mathcal{A}^{cur}$)}
    \STATE $p^{cur} \gets P_1(s^{cur})$ \COMMENT{State validity check}
    \IF{$p^{cur}$}
        \STATE \textbf{return} $r^{cur}$
    \ENDIF
    \STATE $r^{cur} \gets M(cur)$ \COMMENT{Direct Solution Generation}
    \IF{$f(r^{cur})$} 
        \STATE \textbf{return} $r^{cur}$ \COMMENT{$r^{cur}\in{s^{goal}}$ if $r^{cur}$ is correctly verified by $f$}
    \ENDIF

    \STATE $S^{next} \gets \{s^{next}_1, s^{next}_2, ...\}$, $\mathcal{A}^{nc} \gets \{a^{next}_1, a^{next}_2, ...\}$ \COMMENT{Decompose $s^{cur}$ into sub-states $S^{next}$ with correcsponding actions $\mathcal{A}^{nc}$}
    
    \WHILE{$S^{next} \neq \emptyset$}        
        \STATE $S'^{next} \gets \text{Rank}(S^{next})$
        \STATE $s^{next}_i \gets S'^{next}[0]$ \COMMENT{Select most possible state $s^{next}_i$ from $S'^{next}$}
        \STATE $\mathcal{A}^{next} = \mathcal{A}^{cur} + [a^{next}_i]$
        \STATE $r^{next} \gets \text{F}(s^{next}_i, M, f, \mathcal{A}^{next})$
        \IF{$r^{next} \neq None$}
            \STATE \textbf{return} $r^{next}$
        \ENDIF        
        \STATE Remove $s^{next}_i$ from $S^{'next}$
    \ENDWHILE    
    \STATE \textbf{return} None
    
\ENDFUNCTION
\STATE $\mathcal{A} \gets []$
\STATE $\mathcal{A}\gets F(s^{init}, M, F, \mathcal{A})$
\label{algorithm}
\end{algorithmic}
\end{algorithm}

\begin{algorithm}[t!] 
\caption{Algorithm of {\cmethod}.}
\label{alg:Framwork_completeness} 
\begin{algorithmic}
% \REQUIRE $\mathcal{G}=(\mathcal{V}, \mathcal{E}, \mathbf{X})$, $\mathcal{Y}_L$, $\beta$, $T$.
\REQUIRE Initial state $s^{init}$, goal state $s^{goal}$, Language Model $LLM$, external verifier $f$.
\ENSURE Action sequence $\mathcal{A}$ transforming $s^{init}$ to $s^{goal}$

\FUNCTION{F($s^{cur}$, $M$, $f$, $\mathcal{A}^{cur}$)}
    \STATE $r^{cur} \gets M(cur)$ \COMMENT{Direct Solution Generation}
    \IF{$f(r^{cur})$} 
        \STATE \textbf{return} $r^{cur}$ \COMMENT{$r^{cur}\in{s^{goal}}$ if $r^{cur}$ is correctly verified by $f$}
    \ENDIF

    \STATE $S^{next} \gets \{s^{next}_1, s^{next}_2, ...\}$, $\mathcal{A}^{nc} \gets \{a^{next}_1, a^{next}_2, ...\}$ \COMMENT{Learning-guided Complete State Decomposition}
    
    \WHILE{$S^{next} \neq \emptyset$}        
        \STATE $S'^{next} \gets \text{Rank}(S^{next})$ \COMMENT{Two phase state ranking}
        \STATE $s^{next}_i \gets S'^{next}[0]$ \COMMENT{Select most possible state $s^{next}_i$ from $S'^{next}$}
        \STATE $\mathcal{A}^{next} = \mathcal{A}^{cur} + [a^{next}_i]$
        \STATE $r^{next} \gets \text{F}(s^{next}_i, M, f, \mathcal{A}^{next})$
        \IF{$r^{next} \neq None$}
            \STATE \textbf{return} $r^{next}$
        \ENDIF        
        \STATE Remove $s^{next}_i$ from $S^{'next}$
    \ENDWHILE    
    \STATE \textbf{return} None
    
\ENDFUNCTION
\STATE $\mathcal{A} \gets []$
\STATE $\mathcal{A}\gets F(s^{init}, M, F, \mathcal{A})$
\label{algorithm}
\end{algorithmic}
\end{algorithm}

\section{Additional Details of \method and \cmethod}
\label{appendix:additional_details_seal_sealc}
\subsection{Prompt Templates of Direction Solution Generation}
\label{appendix:prompt_templates_direct_solution_generation}
We present the examples of using $p_{solve}$ to directly generate solution for Game of 24 and Mini Crosswords in Table~\ref{tab:direct_solution_generation_game24} and Table~\ref{tab:direct_solution_generation_crosswords}.

\subsection{Prompt Templates of State Validity Checking}
\label{appendix:prompt_templates_state_validity_checking}
We present an example of using $p_{c}$ to check state validity for Game of 24 in Table~\ref{tab:state_validity_checking_prompt}.

\subsection{Prompt Templates of Learning-guided Ranking}
\label{appendix:prompt_templates_learning_guided_ranking}
We present an example of using $p_{v}$ to generate LLM-estimated score for the states in Game of 24 in Table~\ref{tab:learning_guided_ranking_game24}.
\section{Additional Details of Self-Search} 
\label{appendix:more_details_self_search}

To systematically explore the impact of search on LLMs and whether LLMs can develop search capabilities on their own, we consider two types of self-search prompts:
\begin{itemize}[leftmargin=*]
    \item High-level self-search: In this approach, we prompt the LLMs to solve problems through search without specifying which search method to use. This means the LLMs must rely solely on their intrinsic knowledge and reasoning capabilities. We compare this method with \textsc{Vanilla CoT} to assess the importance of search in LLM problem-solving. We show this prompt in Table ~\ref{tab:high_level_self_search_prompt}.
    \item Low-level self-search: In this approach, we explicitly encode the search strategy of \method into the prompt. This method allows us to explore whether LLMs can perform self-search and further demonstrate the effectiveness of \method in improving LLM reasoning for problem-solving. We show this prompt in Table~\ref{tab:low_level_self_search_prompt}.

\end{itemize}

\section{Experimental Settings}
\label{appendix:experimental_settings}
\subsection{Task setup}
\label{appendix:tasks}
In this subsection, we introduce the three tasks used in our experiments. 
To demonstrate the feasibility of \method in various realistic applications, we conduct experiments on  Game of 24~\cite{yao2023tree}, Mini crosswords~\cite{yao2023tree} and Blocksworld~\cite{valmeekam2022large}.

\noindent \textbf{Game of 24}~\cite{yao2023tree}. This task is a mathematical reasoning challenge where the objective is to use four numbers and basic arithmetic operations (+-*/) to obtain the value 24. For example, given the numbers ``$1\ 3\ 3\  7$'', a valid solution is ``$1\times3\times7+3=24$''. The dataset used in this task is publicly available~\footnote{\url{https://www.4nums.com/game/difficulties/}}.
In our experiments, we follow the setup described in~\cite{yao2023tree}, selecting 100 groups of four numbers indexed from 900 to 999 as the target problems. To evaluate the effectiveness of the search methods, we compute the pass rate (PR) over these 100 problems. Additional details regarding the evaluation metrics are provided in Appendix~\ref{appendix:evaluation_metrics}.

\noindent \textbf{Mini Crosswords}~\cite{yao2023tree}. This task, introduced by\cite{yao2023tree}, involves solving a $5 \times 5$ mini crossword puzzle. Specifically, given five horizontal clues and five vertical clues, the aim is to fill a 25-letter grid. We adopt the experimental setup from~\cite{cao2024automating}, wherein each clue is accompanied by a fixed-length list of candidate words. The agent selects words from the list based on the textual descriptions of the clues. The list lengths range from $\{6, 7 , 8 , 9 , 10, 11\}$ words. To generate these word lists, we utilize GPT-4o~\cite{gpt4ocard} to produce candidate words that either share similar meanings with or have more than two matching letters in common with the ground-truth words. Details regarding the prompts used to generate these candidate words are provided in Table~\ref{tab:crosswords_candidate_generation}.

\noindent \textbf{Blocksworld}~\cite{valmeekam2022large}. In this task, the agent must rearrange blocks into specific stacked configurations. A \textbf{state} represents the current arrangement of the blocks, while an \textbf{action} is a command that manipulates them. Each action consists of one of four verbs—\textsc{stack}, \textsc{unstack}, \textsc{put}, or \textsc{pickup}—along with the corresponding objects. The \textbf{action space} is defined by the set of valid actions that comply with the domain constraints and depend on the current state of the blocks.
The agent transitions between states by selecting an action and querying the LLM to predict the resulting changes in the block configuration. The state is then updated by incorporating new conditions for the blocks and removing any conditions that are no longer valid. To evaluate plan correctness, we convert both the generated plans and their corresponding textual problem descriptions into the Planning Domain Definition Language (PDDL)~\cite{silver2024generalized} format. We then use $\textsc{VAL}$ (the Automatic Validation Tool for PDDL)~\cite{howey2003automatic} to assess their validity.

Note that these tasks mainly belong to the offline planning scenario. However, we posit that our method is also applicable to online planning scenarios, where plans are dynamically updated during interactions. We aim to explore the application of our approach to such tasks in future work.

\subsection{Baselines}
\label{appendix:baselines} 
\textbf{Traditional search methods}. 
\begin{itemize}[leftmargin=*]
    \item \textsc{Brute-force (DFS)}: Depth-First Search (DFS) explores each possible solution path by diving deeply into one branch before backtracking to explore others.
    \item \textsc{Brute-force (BFS)}: Breadth-First Search (BFS) systematically explores all intermediate steps at a given depth before progressing to the next level, ensuring all possibilities are considered at each step.
    \item \textsc{Brute-force (DFS-Prune)}: This is a variant of \textsc{DFS}. Specifically, when encountering previously visited states, this method skips their exploration. As a result, it improves the search efficiency of \textsc{Brute-force (DFS)} by reducing the search space.
    \item \textsc{Brute-force (BFS-Prune)}: This is a variant of \textsc{BFS}. Similar to \textsc{Brute-force (DFS-Prune)}, it also skips the exploration of previously visited states, thereby reducing the search space.
\end{itemize}
\textbf{LLM-based search methods}.
\begin{itemize}[leftmargin=*]
    \item \textsc{Best-of-N}~\cite{snell2024scaling}: In this approach, we independently sample $N$ answers from the base LLM and select the best answer according to the Process Reward Model (PRM)'s final answer judgment. Especially, we follow the setting in~\cite{yao2023tree,hao2023reasoning,lightman2023let} to use an LLM as the PRM. 
    \item \textsc{Majority Vote}~\cite{wang2023selfconsistency}: This method is inspired by the intuition that solutions for complex problems are rarely unique. We generate $N=10$ outputs using the LLM and determine the final result by selecting the majority-voted answer among them. This leverages the consensus across multiple outputs to improve accuracy and robustness.
    \item \textsc{Beam search}~\cite{yao2023tree}: Beam search optimizes the PRM by focusing on its per-step predictions to identify the most promising solution paths. Following \cite{snell2024scaling}, we implement a beam search variant based on the BFS-based Tree of Thought (ToT) framework from \cite{yao2023tree}. Specifically, we maintain a beam width  $M=5$ for the three tasks. 
    \item \textsc{Beam search+RV}: 
    While standard LLM-based search methods rely solely on the LLM to obtain the final answers, which is prone to hallucination and scoring misalignment. Inspired by the direct solution generation of \method introduced in Sec.~\ref{sec:seal_method}, we implement \textsc{Beam Search+RV}, which integrates a rule-based verifier into the final decision stage. The framework mirrors standard beam search in its iterative path expansion and scoring but diverges in the terminal step: instead of selecting the answer with the highest PRM score, the rule-based verifier applies deterministic constraints (e.g., task-specific logical checks) to identify admissible solutions. \textsc{Beam search+RV} fails to solve the problem if no candidate satisfies the verification rules.
    % Beam search optimizes the PRM by focusing on its per-step predictions to identify the most promising solution paths. Following \cite{snell2024scaling}, we implement a beam search variant based on the BFS-based Tree of Thought (ToT) framework from \cite{yao2023tree}. Specifically, we maintain a beam width  $M=5$ for the three tasks.
\end{itemize}

\subsection{Language Models}
\label{appendix:llm_setting}
To investigate the feasibility of embodying LLMs into search, we consider several close-sourced and open-sourced LLMs offering state-of-the-art reasoning capabilities: 
\begin{itemize}[leftmargin=*]
    \item \textbf{GPT-4o}~\cite{gpt4ocard}: This model is a widely recognized LLM with strong general-purpose reasoning capabilities, making it a standard LLM for evaluating search methods.
    \item \textbf{GPT-4o-mini}~\cite{achiam2023gpt}: This is a small version of GPT-4o, which has 8B parameters according to~\cite{abacha2024medec}. This model is included to explore the potential of using Small Language Models (SLMs) within \method, particularly to demonstrate its applicability in resource-constrained scenarios.
    \item \textbf{Qwen2.5-72B-Instruct}~\cite{qwen2.5}: This is a representative state-of-the-art open-sourced LLMs developed by Qwen Team. It excels in tasks requiring logical deduction, problem-solving, and multi-step reasoning. The model has been fine-tuned to align with user instructions, making it versatile for various applications, from natural language understanding to complex decision-making tasks.
    \item \textbf{QwQ-32B-Preview}~\cite{qwq-32b-preview}: This is an experimental, state-of-the-art LRM with exceptional logical reasoning and mathematical problem-solving skills. Our experiments reveal that this model explicitly conducts search processes by trying different choices. However, as mentioned in~\cite{qwq-32b-preview}, we also notice that this model may enter circular reasoning patterns, leading to lengthy responses without a conclusive answer. Thus, we involve this model in our experiments to highlight the importance of search in LLM reasoning and the effectiveness of \method's search strategy.
    \item \textbf{DeepSeek-R1}~\cite{deepseekai2025deepseekr1}: The latest open-source LRM, comparable to OpenAI's o1~\cite{openaio1card}. Similar to QwQ-32B-Preview, this model demonstrates explicit search behavior in generating results, further validating the relevance of search methods in improving LLM reasoning.
\end{itemize}

\subsection{Evaluation Metrics}
\label{appendix:evaluation_metrics}
In this subsection, we give the details of the evaluation metrics used in three tasks, which evaluate the performance of the search methods from the perspectives of effectiveness and efficiency.

\noindent \textbf{Pass Rate (PR)}. This metric evaluates the solution quality to evaluate the problem-solving capability of the search methods. The following is how to calculate PR in the three tasks:
\begin{itemize}[leftmargin=*]
    \item Game of 24: PR is calculated as the percentage of generated equations that utilize all the given numbers exactly once and correctly evaluate to 24 across the given set of problems.
    \item Crosswords: PR is evaluated at three levels: letter-level, word-level, and game-level. \textbf{Letter-level PR} measures the proportion of letters in the generated solutions that match exactly with the letters in the ground-truth board. \textbf{Word-level PR} evaluates the percentage of words in the generated solutions that correspond exactly to the ground-truth answers. \textbf{Game-level PR} measures the proportion of problems that are entirely solved, where a problem is considered completely solved only if all letters in the generated solution match the ground-truth answer precisely.
    \item Blockswords: PR is calculated by the percentage of problems where the generated plans successfully achieve the goal within $120\%$ of the minimum required steps.
\end{itemize}

% \noindent\textbf{Search States (SS)}. This metric measures the search methods by counting traversed states in the search space $\mathcal{S}$. 

\noindent\textbf{Search Steps (SS)}. This metric measures the search methods by counting traversed states in the search space $\mathcal{S}$.
To calculate this metric for our \method and \cmethod, it is further broken down into the following components:
\begin{itemize}[leftmargin=*]
    \item LLM Calls, {which is composed of}:
    \begin{itemize}
        \item Number of LLM Answerer Calls: The number of steps taken to directly obtain solutions via LLMs.
        \item Number of State Validity Checking Calls: This refers to the number of steps taken to verify the validity of states. While LLMs are primarily used for state validity checking, we also employ traditional rule-based methods in certain tasks to ensure completeness.
        \item Number of State Ranking Calls: The number of steps taken to rank states.
    \end{itemize}
    \item External Calls, {which is composed of}:
    \begin{itemize}
        \item Number of Decomposition Calls: The number of steps used for state decomposition.
        \item Number of External Verifier Calls: The number of steps used to call the external verifier for validating solutions.
    \end{itemize}
\end{itemize}

For $\textsc{Beam Search}$ and $\textsc{Best-of-N}$, the metric can be broken down into:
\begin{itemize}
    \item LLM Calls:
    \begin{itemize}[leftmargin=*]
    \item Number of LLM Answerer Calls: The number of steps taken to obtain the next-level intermediate steps.
    \item Number of LLM Verifier Calls: The number of steps where the LLM is called as an external verifier for intermediate step evaluation.
\end{itemize}
\end{itemize}

For $\textsc{Majority Voite}$, the metric can be broken down into:
\begin{itemize}
    \item LLM Calls:
    \begin{itemize}[leftmargin=*]
    \item Number of LLM Answerer Calls: The number of steps taken to directly obtain solutions.
    \end{itemize}
\end{itemize}

For the traditional brute-force search methods, $\textsc{DFS}$ and $\textsc{BFS}$, the metric then can be broken down into:
\begin{itemize}
    \item External Calls:
    \begin{itemize}
    \item Number of Decomposition Calls: The number of steps used to perform state decomposition.
    \item Number of Trasversed States: The number of explored states, including both intermediate and final steps.
    \item Number of External Verifier Calls: The number of steps where the external verifier is called to evaluate the final solutions.
\end{itemize}

\end{itemize}
\subsection{Implementation Details}
\label{appendix:implementation_details}
Following the settings in prior studies~\cite{yao2023tree, hao2023reasoning, snell2024scaling}, we set the temperature for LLMs to $\num{0.7}$. To use GPT-4o~\cite{gpt4ocard} and GPT-4o-mini~\cite{achiam2023gpt}, we set the mode as the Chat Completion modes for them. For DeepSeek-R1~\cite{deepseekai2025deepseekr1}, we use the Chat mode to run experiments. 

When using $\textsc{Beam Search}$, we follow the approach from~\cite{yao2023tree} and prompt the LLM three times for each intermediate step, asking whether it was ``sure,'' ``maybe,'' or ``impossible'' to reach 24. We then averaged these responses to evaluate the intermediate steps.

For the baseline search $\textsc{Best-of-N}$, we use the similar setup to $\textsc{Beam Search}$. We evaluate the $N$ solutions at the three levels mentioned above and determine the final result by averaging the responses.

For our \method and \cmethod, when checking the validity of states, such as in the Game of 24, we prompt the LLM five times to get five binary evaluations (``True/False''). The final decision is based on a majority vote to determine whether we should continue exploring or not.

\section{Full Analysis of RQ1: Effectiveness and Efficiency Evaluations}
\label{appendix:rq1_appendix}
To answer \textbf{RQ1}, we conduct experiments on the three tasks using three different LLM backbones: GPT-4o-mini, GPT-4o, and Qwen2.5-72B-Instruct. For Game of 24, we follow~\cite{yao2023tree} to select $100$ problems indexed from $900$ to $999$. For Mini Crosswords, we randomly choose $20$ problems with $11$ candidate words for each clue, while for Blocksworld, we randomly select $20$ problems with a minimum solution length of $8$ steps. The results are reported in Table~\ref{tab:main_result}.
From the table, we observe:
\textbf{(i)} \method achieves a near-perfect pass rate across all settings, outperforming other LLM-based methods, which achieve pass rates of up to 54.0\%, 45.0\%, and 47.4\% for the three tasks, respectively. This validates the effectiveness of \method in problem-solving.
\textbf{(ii)} \method significantly reduces search steps compared to brute-force searches. Specifically, it reduces search steps by up to $99.1\%$ compared to \textsc{DFS} and still achieves state-of-the-art search steps compared to other LLM-based methods. This validates \method's efficiency in navigating the search space.
\textbf{(iii)} \textsc{Beam Search+RV} achieves better pass rates than \textsc{Beam Search}, indicating that integrating rule-based verifiers into LLM-based methods enhances their reliability in problem-solving. However, it still falls behind \method, confirming that current LLM-based methods do not fully conduct real search. This further highlights the superior search effectiveness of \method.
Full search results and the discussion of \method with various verifiers are in Appendix~\ref{appendix:additional_results_seal}.

\section{Additional Experimental Results of \method}
\label{appendix:additional_results_seal}

\subsection{Full Comparison Results on Three Tasks using Various LLMs}
\label{appendix:full_comparison_results_on_three_tasks_three_LLMs}

We present the complete results of the compared search methods using various LLM models for the Game of 24, Mini Crosswords, and Blocksworld tasks in Table~\ref{tab:appendix_full_result_game24}, Table~\ref{tab:appendix_full_result_crosswords}, and Table~\ref{tab:appendix_full_result_blocksworld}, respectively. The results align with the observations discussed in Sec.~\ref{sec:rq1}.

\begin{table*}[t]
\caption{Results of different search methods in the Game of 24 task. }
\centering
\small
\tabcolsep 3.5pt
\renewcommand\arraystretch{1.0}
\begin{tabular}{ll|ccccc}
\toprule
% \multirow{1}{*}{Model} &
% \multirow{1}{*}{Method} &
% \multicolumn{2}{c}{Game of 24} &
% \multicolumn{2}{c}{Mini Crosswords} &
% \multicolumn{2}{c}{Blocksworld} \\
% \cmidrule(r){3-4} \cmidrule(r){5-6} \cmidrule(r){7-8}
\multirow{1}{*}{Model} &
\multirow{1}{*}{Method} & PR (\%) & Avg. SS & Avg. LLM Calls & Avg. Ext. Calls  \\
\midrule
% \multicolumn{8}{l}{\textbf{Traditional Search}} \\
% \cmidrule(r){1-2}
&\textsc{Exhaustive Search}&\num{100} &\num{12928} &\num{0} & \num{12928}\\
% \addlinespace[0.1em]\hdashline\addlinespace[0.1em]
& \textsc{Brute-force (DFS)} & \num{100} & \num{1623} & \num{0} &  \num{1623}  \\

& \textsc{Brute-force (BFS)} & \bfseries\num{100} & \num{3429} & \num{0} & \num{3429} \\
\midrule
% \multicolumn{8}{l}{\textbf{LLM-based Search}} \\
% \cmidrule(r){1-2}
\multirow{5}{*}{\textsc{GPT-4o-mini}}& \textsc{Vanilla CoT} & \num{13} & \bfseries 1 & \num{1} & \num{0}\\

& \textsc{Majority Vote} & \num{15} & \num{10}  & \num{10} & \num{0} \\

& \textsc{Best-of-N} & \num{17} & \num{20}  & \num{20} & \num{0}  \\

& \textsc{Beam Search} & \num{46} & \num{77}& \num{77} & \num{0}  \\
& \textsc{Beam Search+RV} & \num{72} & \num{81}& \num{77} & \num{4}  \\
% \addlinespace[0.1em]\hdashline\addlinespace[0.1em]
% \rowcolor{gray!30} 
& \textsc{\method} &  $\mathbf{100}$ & \num{40.5} & \num{33.4} & \num{7.1} \\
\midrule
% \multicolumn{8}{l}{\textbf{LLM-based Search}} \\
% \cmidrule(r){1-2}
\multirow{5}{*}{\textsc{GPT-4o}}& \textsc{Vanilla CoT} & \num{13} & \bfseries 1  & \num{1} & \num{0}\\

& \textsc{Majority Vote} & \num{23} & \num{10}  & \num{10} & \num{0} \\

& \textsc{Best-of-N} & \num{24} & \num{20}  & \num{20} & \num{0}  \\

& \textsc{Beam Search} & \num{81} & \num{83.4}  & \num{83.4} & \num{0}  \\
& \textsc{Beam Search+RV} & \num{95} & \num{87}& \num{83.4} & \num{3.6}  \\
% \addlinespace[0.1em]\hdashline\addlinespace[0.1em]
% \rowcolor{gray!30} 
& \textsc{\method} &  $\mathbf{100}$ & 66.8 & 55.9 & 10.9 \\
\midrule
% \multicolumn{8}{l}{\textbf{LLM-based Search}} \\
% \cmidrule(r){1-2}
\multirow{5}{*}{\textsc{Qwen2.5-72B}}& \textsc{Vanilla CoT} & \num{17} & \bfseries 1 & \num{1} & \num{0}\\

& \textsc{Majority Vote} & \num{23} & \num{10}  & \num{10} & \num{0} \\

& \textsc{Best-of-N} & \num{27} & \num{20}  & \num{20} & \num{0}  \\

& \textsc{Beam Search} & \num{35} & \num{124} & \num{124} & \num{0}  \\
& \textsc{Beam Search+RV} & \num{70} & \num{128.2}& \num{124} & \num{4.2}  \\
% \addlinespace[0.1em]\hdashline\addlinespace[0.1em]
% \rowcolor{gray!30} 
& \textsc{\method} &  $\mathbf{100}$ & \num{84.9} &  \num{68.5}& \num{16.4} \\
% & \;\; -- \textsc{GPT-4o-mini} & $\mathbf{100}$ & \cellcolor{green!36}\num{40} ({$\downarrow$} 97.5\%) & $\mathbf{100}$ & \cellcolor{green!36}\num{75.8} ({$\downarrow$} 98.2\%) & $\mathbf{100}$ & \cellcolor{green!36}\num{160.6} ({$\downarrow$} 99.1\%)\\

% & \;\; -- \textsc{GPT-4o} & ${99}$ & \cellcolor{green!36}\num{65.6} ({$\downarrow$} 96.0\%) & $\mathbf{100}$ & \cellcolor{green!36}\num{45.0} ({$\downarrow$} 98.9\%) & $\mathbf{100}$ & \cellcolor{green!36}\num{80.8} ({$\downarrow$} 99.5\%) \\

% & \;\; -- \textsc{Qwen2.5-72B-Instruct} & $\mathbf{100}$ & \cellcolor{green!36}\num{84.9} ({$\downarrow$} 94.8\%) & $\mathbf{100}$ & \cellcolor{green!36}\num{98.3} ({$\downarrow$} 98.9\%) & $\mathbf{100}$ & \cellcolor{green!36}\num{68.7} ({$\downarrow$} 99.6\%) \\

\bottomrule
\end{tabular}
\vspace{-5pt}
\label{tab:appendix_full_result_game24}
\end{table*}

\begin{table*}[t]
\caption{Results of different search methods in Mini Crosswords (11-words) task.}
\centering
\small
\tabcolsep 3.5pt
\renewcommand\arraystretch{1.0}
\begin{tabular}{ll|cccccc}
\toprule
Model & Method & Letter-level PR & Word-level PR & Game-level PR & Avg. SS & Avg. LLM Call & Avg. Ext. Call \\
\midrule
% \multicolumn{8}{l}{\textbf{Traditional Search}} \\
% \cmidrule(r){1-2}
&\textsc{Exhaustive Search}&\num{100} &\num{100} &\num{100} &{TL} &\num{0} & {TL}\\
% \addlinespace[0.1em]\hdashline\addlinespace[0.1em]
& \textsc{Brute-force (DFS)} &\num{100}&\num{100}& \num{100} & \num{4128.9} & \num{0} &  \num{4128.9}  \\

& \textsc{Brute-force (BFS)}&\num{100}&\num{100} & \bfseries\num{100} & {TL} & \num{0} & {TL} \\
\midrule
\multirow{5}{*}{\textsc{GPT-4o-mini}} & \textsc{Vanilla CoT} & \num{40.5} & \num{15.9} & \num{0} & \num{1} & \num{1} & \num{0} \\
& \textsc{Majority Vote} & \num{45.0} & \num{19.0} & \num{0} & \num{10} & \num{10.0} & \num{0} \\
& \textsc{Best-of-N} & \num{45.0} & \num{18.5} & \num{0} & \num{20.0} & \num{20.0} & \num{0} \\
& \textsc{Beam Search} & \num{97.8} & \num{90.0} & \num{55.0} & \num{30.9} & \num{30.9} & \num{0} \\
& \textsc{Beam Search+RV} & \num{99.2} & \num{99.0} & \num{90.0} & \num{30.9} & \num{29.1} & \num{1.8} \\
& \textsc{Ours} & \textbf{\num{100}} & \textbf{\num{100}} & \textbf{\num{100}} & \num{75.8} & \num{10.2} & \num{65.6} \\
\midrule
\multirow{5}{*}{\textsc{GPT-4o}} & \textsc{Vanilla CoT} & \num{54.4} & \num{26.0} & \num{5.0} & \num{1} & \num{1} & \num{0} \\
& \textsc{Majority Vote} & \num{61.6} & \num{31.0} & \num{5.0} & \num{10} & \num{10.0} & \num{0} \\
& \textsc{Best-of-N} & \num{65.6} & \num{34.0} & \num{5.0} & \num{20.0} & \num{20.0} & \num{0} \\
& \textsc{Beam Search} & \num{68.0} & \num{61.5} & \num{35.0} & \num{21.3} & \num{21.3} & \num{0} \\
& \textsc{Ours} & \textbf{\num{100}} & \textbf{\num{100}} & \textbf{\num{100}} & \num{45.0} & \num{19.4} & \num{25.6} \\
\midrule
\multirow{5}{*}{\textsc{Qwen2.5-72B}} & \textsc{Vanilla CoT} & \num{58.3} & \num{25.6} & \num{0} & \num{1} & \num{1} & \num{0} \\
& \textsc{Majority Vote} & \num{55.6} & \num{26.3} & \num{5.0} & \num{10} & \num{10.0} & \num{0} \\
& \textsc{Best-of-N} & \num{60.4} & \num{28.5} & \num{5.0} & \num{20} & \num{20.0} & \num{0} \\
& \textsc{Beam Search} & \num{96.8} & \num{88.5} & \num{65.0} & \num{32.0} & \num{32.0} & 0 \\
& \textsc{Ours} & \textbf{\num{100}} & \textbf{\num{100}} & \textbf{\num{100}} & \num{68.1} & \num{27.0} & \num{41.1} \\
\bottomrule
\end{tabular}
\label{tab:appendix_full_result_crosswords}
\end{table*}

\begin{table*}[t]
\caption{Results of different search methods in the Blocksworld (8-steps) task. }
\centering
\small
\tabcolsep 3.5pt
\renewcommand\arraystretch{1.0}
\begin{tabular}{ll|ccccc}
\toprule
\multirow{1}{*}{Model} &
\multirow{1}{*}{Method} & PR (\%) & Avg. SS & Avg. LLM Calls & Avg. Ext. Calls  \\
\midrule
% \multicolumn{8}{l}{\textbf{Traditional Search}} \\
% \cmidrule(r){1-2}
&\textsc{Exhaustive Search}&\num{100} &{TL} &\num{0} & {TL}\\
% \addlinespace[0.1em]\hdashline\addlinespace[0.1em]
& \textsc{Brute-force (DFS)} & \num{100} & \num{18531.9} & \num{0} &  \num{18531.9}  \\

& \textsc{Brute-force (BFS)} & \bfseries\num{100} & \num{96759.4} & \num{0} & \num{96759.4} \\
\midrule
% \multicolumn{8}{l}{\textbf{LLM-based Search}} \\
% \cmidrule(r){1-2}
\multirow{5}{*}{\textsc{GPT-4o-mini}}& \textsc{Vanilla CoT} & \num{0} & \bfseries 1 & \num{1} & \num{0}\\

& \textsc{Majority Vote} & \num{35} & \num{10}  & \num{10} & \num{0} \\

& \textsc{Best-of-N} & \num{0} & \num{20}  & \num{20} & \num{0}  \\

& \textsc{Beam Search} & \num{0} & \num{66.8}& \num{66.8} & \num{0}  \\
& \textsc{Beam Search+RV} & \num{0} & \num{94.8}& \num{90.8} & \num{4.0}  \\
% \addlinespace[0.1em]\hdashline\addlinespace[0.1em]
% \rowcolor{gray!30} 
& \textsc{\method} &  $\mathbf{100}$ & \num{160.6} & \num{50.2} & \num{110.4} \\
\midrule
% \multicolumn{8}{l}{\textbf{LLM-based Search}} \\
% \cmidrule(r){1-2}
\multirow{5}{*}{\textsc{Qwen2.5-72B}}& \textsc{Vanilla CoT} & \num{35} & \bfseries 1 & \num{1} & \num{0}\\

& \textsc{Majority Vote} & \num{60} & \num{10}  & \num{10} & \num{0} \\

& \textsc{Best-of-N} & \num{60} & \num{20}  & \num{20} & \num{0}  \\

& \textsc{Beam Search} & \num{0} & \num{68.7} & \num{68.7} & \num{0}  \\
% \addlinespace[0.1em]\hdashline\addlinespace[0.1em]
% \rowcolor{gray!30} 
& \textsc{\method} &  $\mathbf{100}$ & \num{68.7} &  \num{22.9}& \num{45.8} \\
\bottomrule
\end{tabular}
\vspace{-5pt}
\label{tab:appendix_full_result_blocksworld}
\end{table*}

\subsection{Impact of Verifier in Search with \method}
\label{appendix:impact_verifier_search}
In Sec.~\ref{sec:seal_method}, we introduce the direct solution generation component of \method, which leverages LLMs to generate solutions and uses a verifier \( f \) to evaluate their correctness. The verifier can take various forms, such as a traditional rule-based verifier, a PRM~\cite{lightman2023let}, or an ORM~\cite{uesato2022solving}. For our implementation, we opt for a rule-based verifier such that we can focus on explorating search. 
However, there are scenarios where a verifier \( f \) may not be available. To evaluate the effectiveness of \method under such conditions, we implement a variant of \method that excludes both the verifier and the direct solution generation component. The results on the \textit{Game of 24} task using GPT-4o-mini are shown in Table~\ref{tab:impact_of_verifier}. 
From the table, we observe that both variants of \method—with and without a verifier—achieve high accuracy. Notably, the variant without a verifier (\method+NV) achieves a \num{90}\% pass rate while requiring only \num{44.9} search steps. This demonstrates the effectiveness and flexibility of \method across various settings. 

\begin{table*}[t]
\caption{Result of the impact of verifier in the Game of 24 task using GPT-4o-mini.}
\centering
\small
\tabcolsep 3.5pt
\renewcommand\arraystretch{1.0}
\begin{tabular}{ll|ccccc}
\toprule
\multirow{1}{*}{Method} & PR (\%) & Avg. SS & Avg. LLM Calls & Avg. Ext. Calls  \\
\midrule
\textsc{Vanilla CoT} & \num{0} & \bfseries 1 & \num{1} & \num{0}\\

\textsc{Majority Vote} & \num{35} & \num{10}  & \num{10} & \num{0} \\

\textsc{Best-of-N} & \num{0} & \num{20}  & \num{20} & \num{0}  \\
\textsc{Beam Search} & \num{46} & \num{77}& \num{77} & \num{0}  \\
\textsc{Beam Search+RV} & \num{46} & \num{77}& \num{77} & \num{0}  \\
% \textsc{\method (LLM verifier)} & \num{69} & \num{77}& \num{77} & \num{0}  \\

\textsc{\method}+NV & \num{90} & \num{44.9}& \num{38.2} & \num{6.7}  \\

\textsc{\method} &  $\mathbf{100}$ & \num{40.5} & \num{33.4} & \num{7.1} \\
\bottomrule
\end{tabular}
\vspace{-5pt}
\label{tab:impact_of_verifier}
\end{table*}

\section{Additional Experimental Results of \cmethod}
\label{appendix:addtional_impact_difficulty_of_seal_c}
\begin{figure*}[t]
    \small
    \centering
    \begin{subfigure}{0.24\textwidth}
        \includegraphics[width=0.98\linewidth]{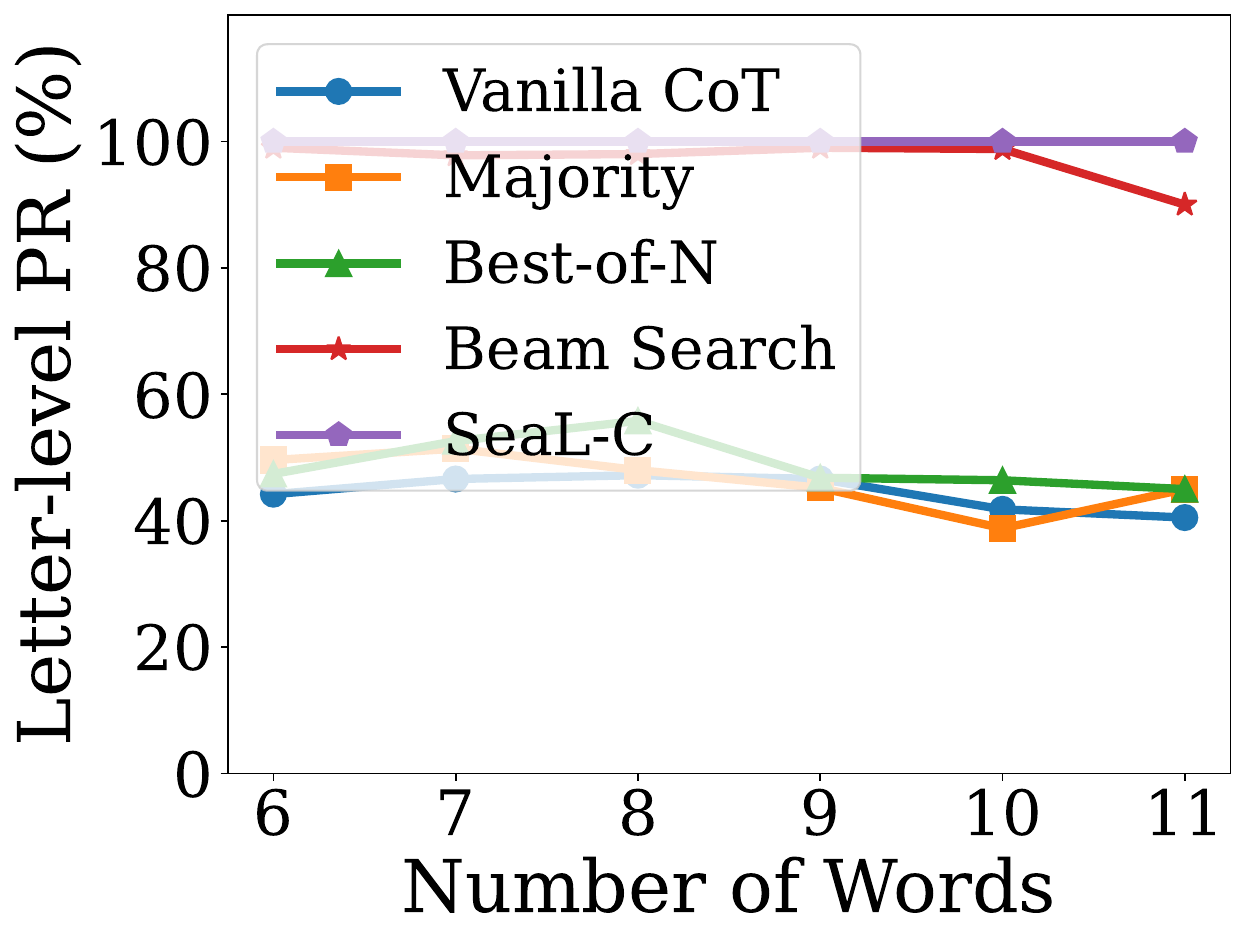}
        \vskip -0.5em
        \caption{Letter-level PR}
    \end{subfigure}
    \begin{subfigure}{0.24\textwidth}
        \includegraphics[width=0.98\linewidth]{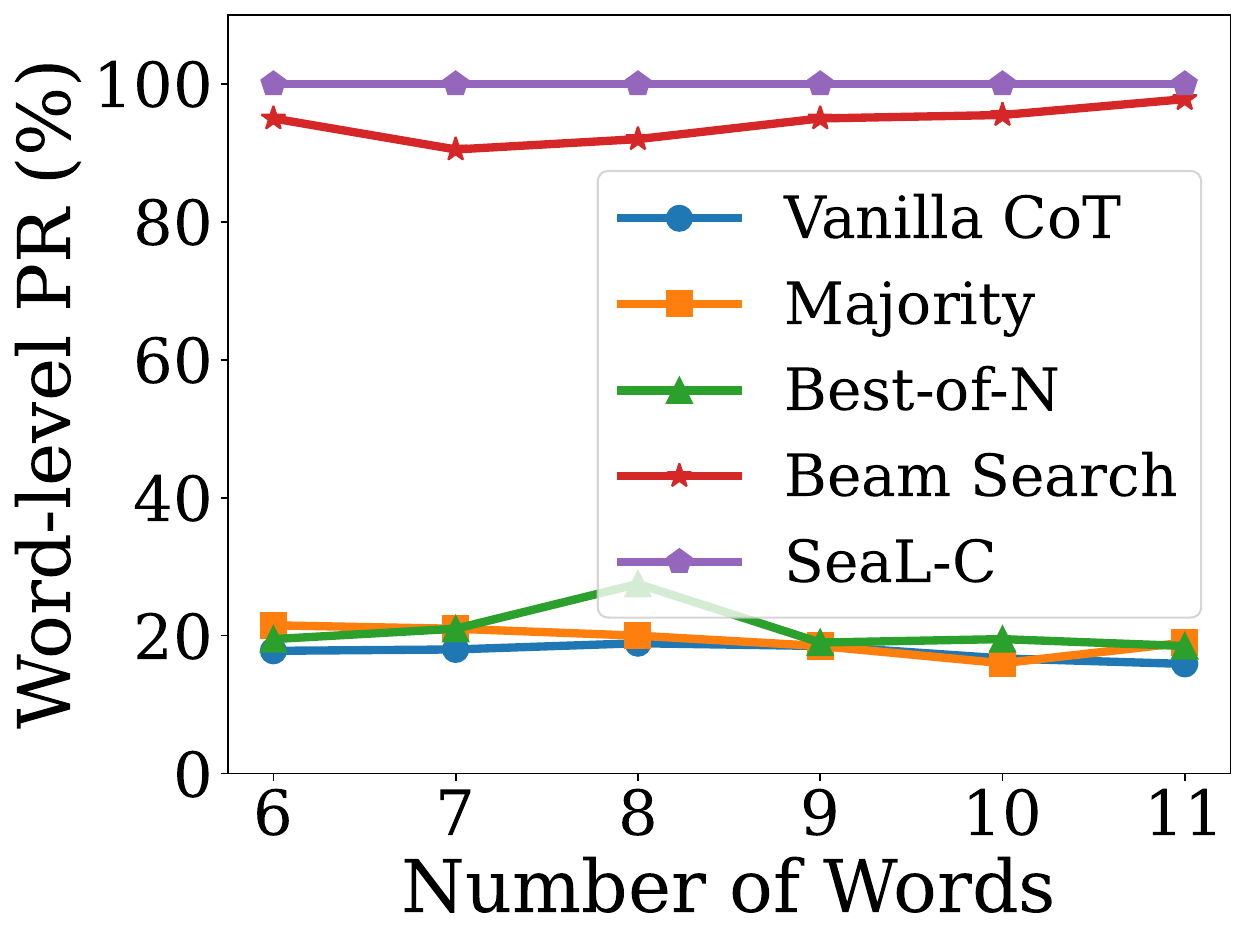}
        \vskip -0.5em
        \caption{Word-level PR}
    \end{subfigure}
    \begin{subfigure}{0.24\textwidth}
        \includegraphics[width=0.98\linewidth]{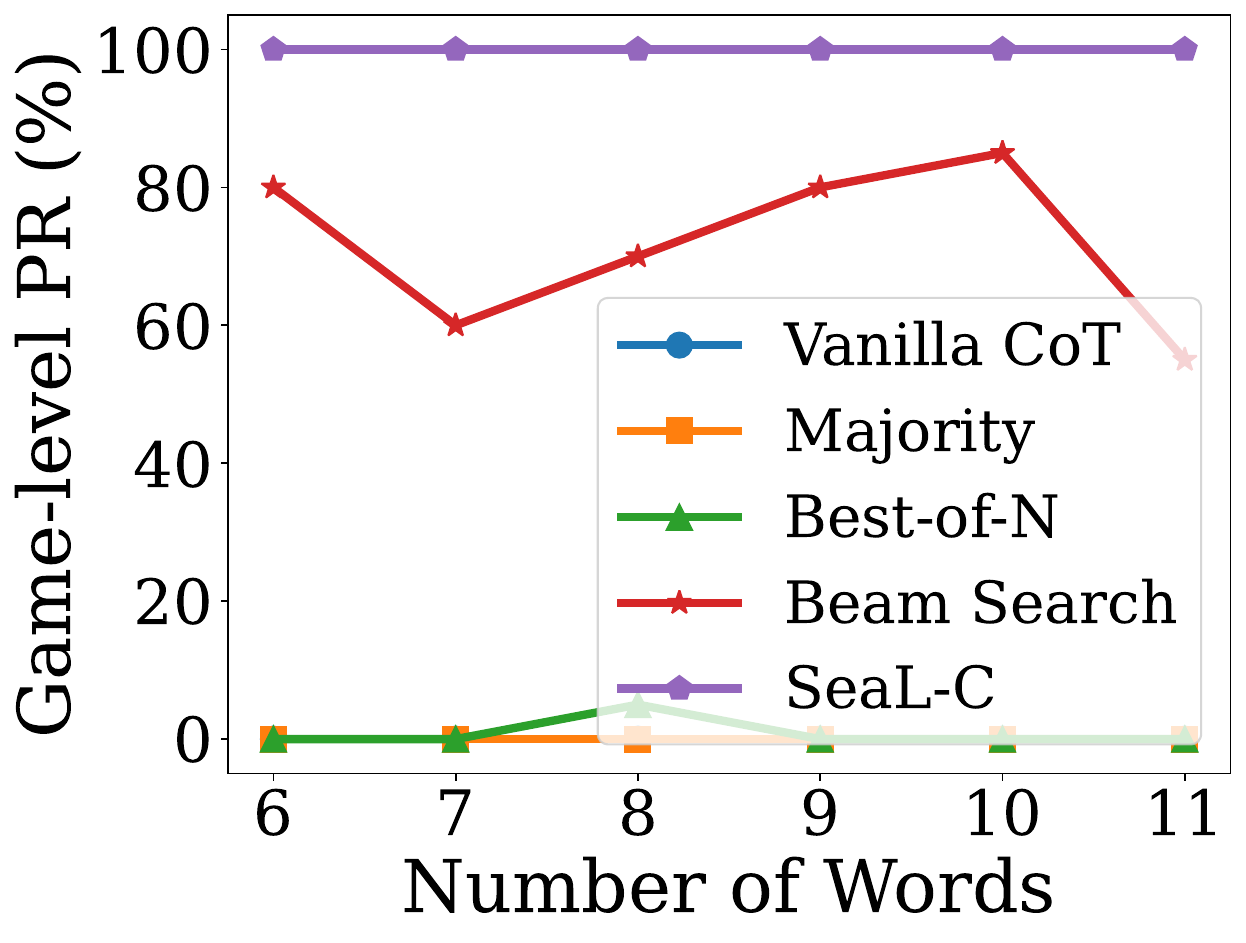}
        \vskip -0.5em
        \caption{Game-level PR}
    \end{subfigure}
    \begin{subfigure}{0.24\textwidth}
        \includegraphics[width=0.98\linewidth]{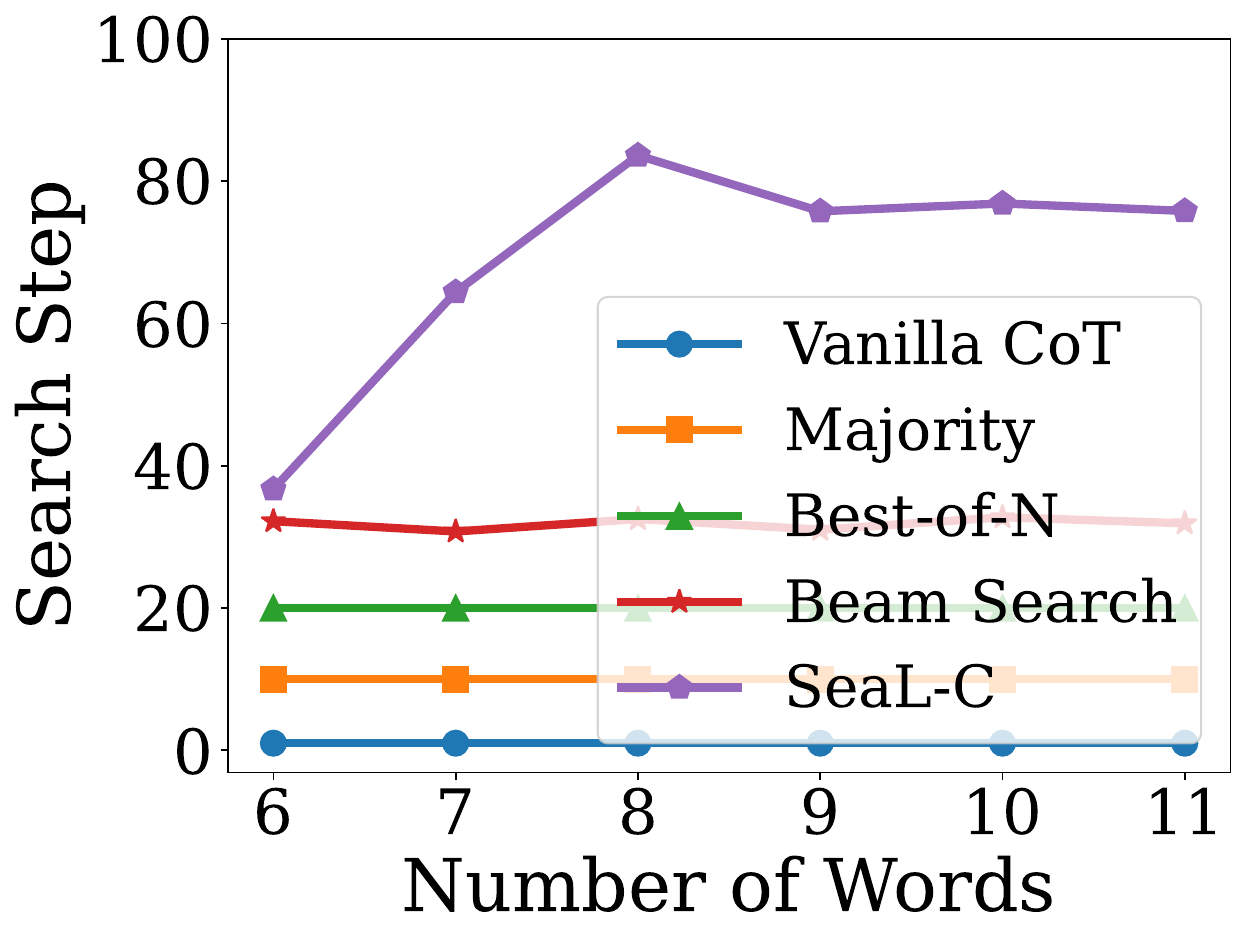}
        \vskip -0.5em
        \caption{SS}
    \end{subfigure}
    \vskip -1.em
    \caption{Impact of problem difficulty on the completeness of search on Game of 24 and Blocksworld tasks using GPT-4o-mini.}
    \vskip -1em
    \label{fig:impact_problem_difficulty_completeness_minicrosswords_appendix}
\end{figure*}

In this section, we will provide the additional results about the impact of problem difficulty on the completeness of search on Mini Crosswords task using GPT-4o-mini, which is shown in Fig~\ref{fig:impact_problem_difficulty_completeness_minicrosswords_appendix}. Specifically, Fig.~\ref{fig:impact_problem_difficulty_completeness_minicrosswords_appendix} (a) - (c) present the results of letter-level, word-level and game-level pass rates, respectively. Fig.~\ref{fig:impact_problem_difficulty_completeness_minicrosswords_appendix} (d) is the result of search step. The results align with the observations discussed in Sec.~\ref{sec:rq2}.

\begin{figure}[t]
    \small
    \centering
    
        \includegraphics[width=0.8\linewidth]{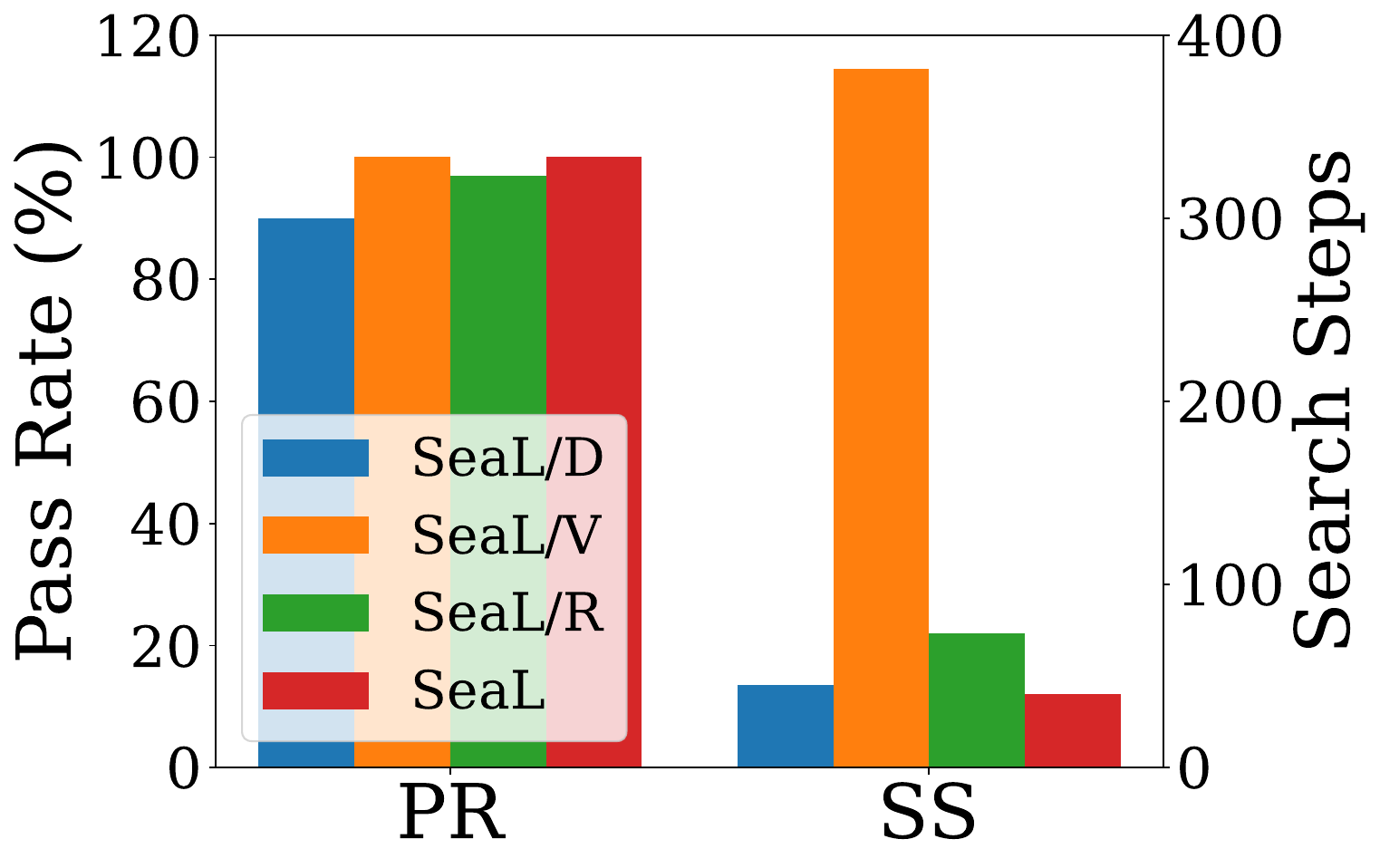}
        \vskip -0.5em

    \vskip -1.em
    \caption{Ablation studies on Game of 24 using GPT-4o-mini to explore the contribution of each key component in \method.}
    \vskip -1em
    \label{fig:ablation_studies_components_24game_gpt4omini_appendix}
\end{figure}

\section{Additional Details of Problem-solving Processes of LRMs}
\label{appendix:more_details_of_problem_solving_processes_LRMs}
In this section, we present a detailed analysis of how existing Large Retrieval Models (LRMs) utilize search strategies to solve problems, underscoring the critical role of search in effective problem-solving. 

Table~\ref{tab:qwq_LRM_game24} illustrates an example of QwQ-32B-Preview's problem-solving process in the Game of 24 task. From this table, we observe that QwQ-32B-Preview explicitly explores multiple potential answers through an iterative search process, continuing until a correct solution is identified and verified. This mechanism of LRMs likely explains why LRMs exhibit superior problem-solving capabilities compared to general LLMs such as GPT-4o, further emphasizing the importance of search in leveraging LLMs for complex problem-solving.

However, we also note a limitation in QwQ-32B-Preview’s approach: its search strategy sequentially evaluates and verifies candidate solutions one by one, which may be suboptimal in terms of efficiency. In contrast, Tables~\ref{tab:qwq_LRM_game24_seal1},\ref{tab:qwq_LRM_game24_seal2}, and\ref{tab:qwq_LRM_game24_seal3} illustrate the problem-solving process when employing our \method's search strategy. From the table, we observe that QwQ-32B-Preview can understand and execute our \method's search strategy, allowing it to arrive at solutions more efficiently and effectively—without relying on exhaustive trial-and-error. This observation suggests that LRMs can significantly benefit from more advanced search methodologies, reinforcing the need to explore ways to enhance the self-search capabilities of LLMs.

\section{Additional Experimental Results of Ablation Studies}
\label{appendix:additional_results_of_ablation_studies}
In this section, we present additional ablation studies to analyze the impact of each key component of \method, as introduced in Sec.~\ref{sec:seal_method}, and to further explore their contributions toward ensuring effective and efficient search. Specifically, we evaluate several ablated variants: (i) \method/V: this variant removes the state validity checking and directly proceeds to rank the substates after decomposition; (ii) \method/D: this variant excludes the direct solution generation component, relying solely on the generation of next-level intermediate steps at each stage; and (iii) \method/R: this variant disregards the learning-guided ranking component, exploring substates sequentially based on their default order. We compare \method against these variants on the Game of 24 using GPT-4o-mini, and the results are summarized in Fig.~\ref{fig:ablation_studies_components_24game_gpt4omini_appendix}. From the figure, we observe the following:
\begin{itemize}[leftmargin=*]
    \item While \method achieves comparable pass rates to \method/V, it requires significantly fewer search steps. This improvement is attributed to the state validity checking component, which leverages the reasoning capabilities of LLMs to evaluate state validity and determine whether further exploration is warranted. By filtering out invalid states, the search budget is conserved, preventing unnecessary exploration. These findings corroborate our earlier analysis in Sec.~\ref{sec:preliminary_results_analysis}.
    \item Both \method/D and \method achieve similar pass rates; however, \method/D requires more search steps. This discrepancy arises because the direct solution generation component in \method can resolve simpler problems within intermediate substates in a single step, eliminating the need for further exploration. This observation aligns with our analysis in Sec.~\ref{sec:preliminary_results_analysis}, which highlights the capability of LLMs to solve simpler problems more efficiently.
    \item The search steps for \method/R are higher than for \method, confirming the effectiveness of the learning-guided ranking component in prioritizing valid states that lead to the goal state earlier. Notably, \method/R achieves a slightly lower yet comparable pass rate to \method. This reduction can be attributed to occasional misclassification of valid states as invalid due to the limitations in the reasoning capabilities of LLMs. This result underscores the necessity of incorporating the full \method pipeline in scenarios where a $100\%$ pass rate is critical, particularly in high-stakes applications.
\end{itemize}

\begin{table*}[ht]
\small
\centering
\caption{The prompt of direct solution generation for solving Game of 24}
\begin{tabular}{p{0.9\linewidth}} \toprule
Use the given numbers and basic arithmetic operations (+, -, , /) to reach 24. Each step should only select two of the remaining numbers to calculate a new number, aiming to reduce the total count of numbers by merging the selected pair into their result. The steps should methodically progress towards constructing an expression that equals 24 when evaluated. Each remaining number can be only selected up to once. For example:\\
\\
Input: 4 4 6 8\\
Steps:\\
step1: 4 + 8 = 12; (left: 4 6 12)\\
step2: 6 - 4 = 2; (left: 2 12)\\
step3: 2 * 12 = 24; (left: 24)\\
Answer: (6 - 4) * (4 + 8) = 24\\
\\
\textcolor{blue}{\textsc{[more few-show examples]}}\\
Ensure that each arithmetic operation is possible and leads to the correct remaining numbers. The final answer should correctly reflect the steps performed to achieve 24. If an error occurs in calculation, revise the final expression accordingly. \\
Given the following input, the generated output should be formatted exactly as above: \\
Input: \textcolor{blue}{\textsc{[input]}}
\\ \bottomrule
\end{tabular}
\label{tab:direct_solution_generation_game24}
\end{table*}

\begin{table*}[ht]
\small
\centering
\caption{The prompt of direct solution generation for solving Mini Crosswords}
\begin{tabular}{p{0.9\linewidth}} \toprule
Solve 5x5 mini crosswords by selecting appropriate words from provided candidate lists. Given an input of 5 horizontal clues and 5 vertical clues, a list of words is given for each clue. \\
Consider intersecting constraints with other words, solve this step by step, generate thoughts about 5-letter word from the corresponding list fits each clue, and then select the most suitable word for each clue.\\
Then an output of 5 rows, where each row is 5 letter separated by space.\\
\\
\# Few-shot example 1\\
Input:\\
h1. A lunar valley\\
h2. A fatty oil\\
h3. To entice\\
h4. To lower; to reduce\\
h5. A solitary person\\
v1. According to the roster\\
v2. Another name for Port-Francqui\\
v3. An illicit lover; a European lake\\
v4. To lisp\\
v5. To come in\\
\\
Thoughts:\\
h1. A lunar valley: RILLE\\
h2. A fatty oil: OLEIN\\
h3. To entice: TEMPT\\
h4. To lower; to reduce: ABASE\\
h5. A solitary person: LONER\\
v1. According to the roster: ROTAL\\
v2. Another name for Port-Francqui: ILEBO\\
v3. An illicit lover; a European lake: LEMAN\\
v4. To lisp: LIPSE\\
v5. To come in: ENTER\\
\\
Output:\\
R I L L E\\
O L E I N\\
T E M P T\\
A B A S E\\
L O N E R\\
\\
Input:\\
\textcolor{blue}{\textsc{input}}
\\ \bottomrule
\end{tabular}
\label{tab:direct_solution_generation_crosswords}
\end{table*}

\begin{table*}[ht]
\small
\centering
\caption{The prompt of state validity checking for solving Game of 24}
\begin{tabular}{p{0.9\linewidth}} \toprule
Your task is to analyze a mathematical game where the goal is to use basic arithmetic operations (+, -, *, /) to achieve the target number 24. You are given several sets of numbers. For each set, determine if it is possible (likely/unlikely) to achieve 24 using any combination of numbers and operations. For each set, each number must be used exactly once. If one of these sets are likely to obtain 24, return the answer Yes and these sets that are likely to reach to 24.\\
\\
Examples:\\
Example 1:\\
Input:\\
State 0: 8 3 1\\
State 1: 6 4\\
State 2: 7 7 2\\
\\
State Precheck:\\
Answer: yes; Reason: State 1 is likely to reach 24\\
\\
Example 2:\\
Input:\\
State 0: 8 \\
State 1: 11 2.66\\
State 2: 7 7 \\
\\
State Precheck:\\
Answer: no;\\
\\
Example 3:\\
Input:\\
State 0: 24\\
State 1: 12\\
\\
State Precheck:\\
Answer: yes; Reason: State 0 and State 2 can directly reach to 24 because the single number 24 is directly equal to 24, 24 = 24.\\
\\
Given the following input sets, generate the output in the exact format above: \\
Input:\\
\textcolor{blue}{\textsc{[input]}}
\\ \bottomrule
\end{tabular}
\label{tab:state_validity_checking_prompt}
\end{table*}

\begin{table*}[ht]
\small
\centering
\caption{The prompt of learning-guided state ranking for solving Game of 24}
\begin{tabular}{p{0.9\linewidth}} \toprule
Evaluate if given numbers can reach 24, each given number must be used exactly once (sure/likely/impossible)\\
Current numbers: 10 14; Calculation: 10 + 14 = 24; Comment: I can obtain the 24 by using current numbers; Conclusion: sure\\
Current numbers: 11 12; Calculation: 11 + 12 = 23; Calculation: 12 - 11 = 1; Calculation: 11 * 12 = 132; Calculation: 11 / 12 = 0.91; Comment: I cannot obtain the 24 by using current numbers; Conclusion: impossible\\
Current numbers: 5 7 8; Calculation: 5 + 7 + 8 = 12 + 8 = 20; Calculation: (8 - 5) * 7 = 3 * 7 = 21; Comment: I cannot obtain 24 now, but numbers are within a reasonable range; Conclusion: likely\\
\textcolor{blue}{[\textsc{more few-shot examples}]}
\\
Given the following current number, the generated output should be formatted exactly as above: 
Current number: {input};
\\ \bottomrule
\end{tabular}
\label{tab:learning_guided_ranking_game24}
\end{table*}

\begin{table*}[ht]
\small
\centering
\caption{The prompt of high-level self-search for solving Game of 24}
\begin{tabular}{p{0.9\linewidth}} \toprule
% \textbf{System Prompt} 
% \colorbox{lightgreen}{generally decreasing trend,}
% \colorbox{yellow}{This part has a yellow background}

You are a problem solver for the Game of 24. Given four numbers, your goal is to find a mathematical expression that equals 24 using each number exactly once, with the allowed operations being addition (+), subtraction (-), multiplication (*), and division (/). Follow these steps precisely:\\
\\
\# Game of 24 Rules\\
Rule 1. Each number must be used exactly once.\\
Rule 2. Use only basic operation (+-*/) for calculation.\\
\\
\# Input\\
You will receive four numbers. For example: "Input: 1 2 4 7"\\
\\
\# Instruction:\\
You are required to get 24 with the 4 input numbers by using the idea of search for your thoughts in the intermediate steps.\\
\\
Wrap your your final solution in a special tag <solution> like <solution> 2 + 3 * 5 + 7 = 24 </solution>. Stop when find a correct solution.\\
Input: \textcolor{blue}{\textsc{[input]}}
\\ \bottomrule
\end{tabular}
\label{tab:high_level_self_search_prompt}
\end{table*}

\begin{table*}[t]
\small
\centering
\caption{The prompt of low-level self-search for solving Game of 24}
\begin{tabular}{p{0.9\linewidth}} \toprule
You are a problem solver for the Game of 24. Given four numbers, your goal is to find a mathematical expression that equals 24 using each number exactly once, with the allowed operations being addition (+), subtraction (-), multiplication (*), and division (/). Follow these steps precisely:\\
\# Game of 24 Rules\\
Rule 1. Each number must be used exactly once.\\
Rule 2. Use only basic operation (+-*/) for calculation.\\
Rule 3. Not to generate code to solve the game. And not to return empty content.\\
\\
\# Input\\
You will receive four numbers. For example: "Input: 1 2 4 7"\\
\\
\# Instruction\\
You are required to get 24 with the 4 input numbers by using the idea of search to formalize your thoughts in the intermediate steps. Specifically, the idea search is based on the following stages:\\
Stage 1. State Precheck. Precheck if it is possible to obtain 24 from the numbers in the current state, where each number is used exactly once. Stop expansion in the current state if precheck fails.\\
Stage 2. Direct Solution Attempt. Attempt to directly find a solution using the current numbers by thinking step by step. \\
Stage 3. Check the the correctness of the generated direct solution. If correct, stop early and use it as the final solution. If not, jump to the next stage. Note that all numbers in the solution must from current state and be used exactly once.\\
Stage 4. Problem decompositon. Decompose current problem into several sub-problems by using basic operation (+-*/) to see if you can solve these subproblems to get 24. For instance. "1 2 4 7" can be decomposed into "2 4 8" by "1 + 7" and "1 2 7" by "4 / 2", where thinking about "2 4 8" is easier than " 1 2 4 7".\\
Stage 5. State ranks. After getting several decomposed sub-problems, you will evaluate each substate's potential to reach 24. The metric is sure = 1.0, likely = 0.5, impossible = 0.1. Then give a sorted substates list, highest potential first.\\
\\
Repeat the above stages until a valid action is found. Note that you can conduct backtrace to switch to another subproblem if current subproblem cannot achieve 24. For example, if you find that "1 2 7" cannot achieve 24 even by further decomposition, like "2 7", "1 7" from "1 2 7" or even "14" from "2 7", you can backtrace to another high-layer subproblems like "2 4 8" until you finally find a solution. \\
\\
Wrap your your final solution in a special tag <solution> like <solution> 2 + 3 * 5 + 7 = 24 </solution>. Stop when find a correct solution.\\
Input: \textcolor{blue}{\textsc{[input]}}
\\ \bottomrule
\end{tabular}
\label{tab:low_level_self_search_prompt}
\end{table*}

\begin{table*}[ht]
\small
\centering
\caption{The prompt of generating candidate words for clues in the Mini Crosswords task}
\begin{tabular}{p{0.9\linewidth}} \toprule
Generate different types of 5-letter words based on a clue and meaning:\\
\\
Clue: "\textcolor{blue}{\textsc{[textual clue]}}"\\
Ground truth word: \textcolor{blue}{\textsc{[ground-truth word]}}\\
\\
Generate EXACTLY:\\
1. \textcolor{blue}{\textsc{[number of words with similar meainings]}} five-letter words that are semantically related to the clue's meaning (different from ground truth)\\
2. \textcolor{blue}{\textsc{[number of words with similar letters]}} five-letter words that share at least 2 letters in the same positions as the ground truth word "\textcolor{blue}{\textsc{[ground-truth word]}}"\\
\\
Requirements:\\
- Each word must be exactly 5 letters\\
- Do not include the ground truth word\\
- Related words must have meanings close to the clue\\
- Letter-matching words must share at least 2 letters in the same positions as the ground truth word \\
- Generate exactly the requested number of words\\
\\
Format your response strictly as:\\
Related words (\textcolor{blue}{\textsc{[number of words with similar meainings]}} words):\\
- word1: brief explanation of relevance\\
- word2: brief explanation of relevance\\

[continue until exactly \textcolor{blue}{\textsc{[number of words with similar meainings]}} words]\\
\\
Unrelated words (\textcolor{blue}{\textsc{[number of words with similar letters]}} words):\\
- word1\\
- word2\\
 
<continue until exactly \textcolor{blue}{\textsc{[number of words with similar letters]}} words>
\\\bottomrule
\end{tabular}
\label{tab:crosswords_candidate_generation}
\end{table*}

\begin{table*}[ht]
\small
\centering
\caption{An example of the problem-solving process of using QwQ-32B-Preview for the Game of 24 task}
\begin{tabular}{p{0.9\linewidth}} \toprule
\textbf{Input: }\\
$4\ 5\ 6\ 10$\\
\midrule
\textbf{Problem-solving process:}\\
To solve the game of 24 with the numbers 4, 5, 6, and 10, we need to form an expression using each number exactly once and the operations of addition (+), subtraction (-), multiplication (*), and division (/) to reach the result of 24.\\
\\
Let's explore different combinations step by step.\\
\\
First, let's consider multiplication and addition since they are likely to help us reach higher numbers.\\
\\
Option 1:\\
Multiply 4 and 6 to get 24, but we still need to incorporate 5 and 10. However, since we have to use all numbers exactly once, this direct multiplication alone isn't sufficient.\\
\\
Option 2:\\
Consider combining addition and multiplication.\\
\\
Let's try:\\
\\
(5 - 4) * 10 + 6 = 1 * 10 + 6 = 10 + 6 = 16
\\
Not 24.\\
\textcolor{red}{\textsc{[Omitting 10 additional options....]}}\\
\\
Option 13:\\
Another approach:\\
\\
(5 + 10) / 5 * 4 = 15 / 5 * 4 = 3 * 4 = 12\\
\\
Not 24.\\
Option 14:\\
Let's try:\\
\\
(10 - 4) * 5 - 6 = 6 * 5 - 6 = 30 - 6 = 24\\
\\
Yes!\\
\\
So, one possible solution is:\\
\\
(10 - 4) * 5 - 6 = 24\\
\\
Let's verify:\\
\\
First, subtract 4 from 10: 10 - 4 = 6\\
\\
Then, multiply by 5: 6 * 5 = 30\\
\\
Finally, subtract 6: 30 - 6 = 24\\

Perfect!
\\ \bottomrule
\end{tabular}
\label{tab:qwq_LRM_game24}
\end{table*}

\begin{table*}[ht]
\small
\centering
\caption{An example of the problem-solving process of using \textsc{Self-Search (Low)} QwQ-32B-Preview for the Game of 24 task.}
\begin{tabular}{p{0.9\linewidth}} \toprule
\textbf{Input: }\\
$3\ 3\ 7\ 12$\\
\midrule
\textbf{Problem-solving process:}\\
I'm going to try to solve this Game of 24 puzzle with the numbers 3, 3, 7, and 12. The goal is to use each number exactly once with the operations of addition, subtraction, multiplication, and division to reach 24.\\
\\
**Stage 1: State Precheck**\\
\\
First, I need to check if it's possible to get 24 with these numbers. All numbers are positive integers, and 24 is also a positive integer, so it's plausible. I should consider the range of possible values. The smallest value I can get might be around the smallest number, which is 3, and the largest could be much higher, especially with multiplication. So, it seems possible.\\
\\
**Stage 2: Direct Solution Attempt**\\
\\
Let me try to find a direct solution by trying out some combinations:\\
\\
- 3 + 3 + 7 + 12 = 25 (too high)\\
\\
- 12 * 2 = 24, but I don't have a 2 directly. Maybe (7 - 3) = 4, but 12 * 4 = 48, which is too high.\\
\\
- 7 * 3 = 21, then 21 + 3 = 24, but I still have 12 left to use, which is a problem because I have to use each number exactly once.\\
\\
- Wait, 21 + 3 is 24, but I have an extra 12. Maybe I can involve division or subtraction.\\
\\
- What if I do 21 + (12 / 3) = 21 + 4 = 25, which is too high.\\
\\
- Alternatively, 21 + 3 - 12 = 12, which is too low.\\
\\
- Maybe another approach: 12 * 7 = 84, then 84 / 3 = 28, which is too high.\\
\\
- 84 / 3.5, but I don't have 3.5; I have two 3's and a 7 and 12.\\
\\
- Perhaps I can do 12 * (7 - 3) = 12 * 4 = 48, which is too high.\\
\\
- Then 48 / 2, but again, I don't have a 2.\\
\\
- Wait, I have two 3's, so maybe 3 / 3 = 1, but 48 / 1 = 48, still too high.\\
\\
- Maybe another path: 7 * 12 = 84, then 84 / 3 = 28, which is still too high.\\
\\
- 28 - 4 = 24, but I don't have a 4.\\
\\
- Hmm.\\
\\
**Stage 3: Check Direct Solution**
\\
From the above attempts, I haven't found a direct solution yet. So, I need to proceed to decompose the problem.\\
\\ \bottomrule
\end{tabular}
\label{tab:qwq_LRM_game24_seal1}
\end{table*}

\begin{table*}[ht]
\small
\centering
\caption{An example of the problem-solving process using \textsc{Self-Search (Low)} with QwQ-32B-Preview for the Game of 24 task (continued from Tab.~\ref{tab:qwq_LRM_game24_seal1}).}
\begin{tabular}{p{0.9\linewidth}} \toprule
\textbf{Input: }\\
$3\ 3\ 7\ 12$\\
\midrule
\textbf{Problem-solving process:}\\
**Stage 4: Problem Decomposition**\\
\\
Let me try to decompose the problem into sub-problems. I'll pick two numbers and combine them using one of the operations, then see if I can reach 24 with the resulting number and the remaining two numbers.\\
\\
Option 1: Combine 3 and 3.\\
\\
- 3 + 3 = 6\\
\\
- Now, the remaining numbers are 6, 7, 12.\\
\\
- Can I get 24 from 6, 7, 12?\\
\\
- 6 * 7 = 42, then 42 - 12 = 30, too high.\\
\\
- 6 * 12 = 72, then 72 / 7 = 10.285, not 24.\\
\\
- 7 * 12 = 84, then 84 / 6 = 14, not 24.\\
\\
- 12 - 6 = 6, then 6 * 7 = 42, which is too high.\\
\\
- 12 + 7 = 19, then 19 * 6 = 114, too high.\\
\\
- Doesn't seem promising.\\
\\
Option 2: Combine 3 and 7.\\
\\
- 3 + 7 = 10\\
\\
- Remaining numbers: 10, 3, 12.\\
\\
- Can I get 24 from 10, 3, 12?\\
\\
- 10 + 12 = 22, then 22 + 3 = 25, too high.\\
\\
- 10 * 12 = 120, then 120 / 3 = 40, too high.\\
\\
- 12 - 10 = 2, then 2 * 3 = 6, too low.\\
\\
- 12 / 10 = 1.2, then 1.2 * 3 = 3.6, too low.\\
\\
- Doesn't seem to work.\\
\\
Option 3: Combine 3 and 12.\\
\\
- 3 + 12 = 15\\
\\
- Remaining numbers: 15, 3, 7.\\
\\
- Can I get 24 from 15, 3, 7?\\
\\
- 15 + 7 = 22, then 22 + 3 = 25, too high.\\
\\
- 15 * 7 = 105, then 105 / 3 = 35, too high.\\
\\ \bottomrule
\end{tabular}
\label{tab:qwq_LRM_game24_seal2}
\end{table*}

\begin{table*}[ht]
\small
\centering
\caption{An example of the problem-solving process using \textsc{Self-Search (Low)} with QwQ-32B-Preview for the Game of 24 task (continued from Tab.~\ref{tab:qwq_LRM_game24_seal2}).}
\begin{tabular}{p{0.9\linewidth}} \toprule
\textbf{Input: }\\
$3\ 3\ 7\ 12$\\
\midrule
\textbf{Problem-solving process:}\\
- 15 - 7 = 8, then 8 * 3 = 24. Wait, that's 24!\\
\\
- So, 15 - 7 = 8, then 8 * 3 = 24.\\
\\
- But I need to use all numbers exactly once, so let's see:\\
\\
- Original numbers: 3, 3, 7, 12.\\
\\
- Combined 3 and 12 to get 15.\\
\\
- Then, 15 - 7 = 8.\\
\\
- Then, 8 * 3 = 24.\\
\\
- Wait, but I've used one 3 in the combination and another 3 separately, which is fine since there are two 3's.\\
\\
- So, this seems to work.\\
\\\
**Stage 5: State Ranks**\\
\\
I've found a potential solution: (3 + 12 - 7) * 3 = 24.\\
\\
Let me verify:\\
\\
- 3 + 12 = 15\\
\\
- 15 - 7 = 8\\
\\
- 8 * 3 = 24\\
\\
Yes, that's correct.\\
\\
**Final Solution**\\
\\
\[ (3 + 12 - 7) \times 3 = 24 \]\\
\\ \bottomrule
\end{tabular}
\label{tab:qwq_LRM_game24_seal3}
\end{table*}

\end{document}